\documentclass[sigconf,edbt]{acmart-edbt2020}

\def\BibTeX{{\rm B\kern-.05em{\sc i\kern-.025em b}\kern-.08em
    T\kern-.1667em\lower.7ex\hbox{E}\kern-.125emX}}

\usepackage{booktabs} 
\usepackage{graphicx}
\usepackage{footnote}
\usepackage{caption}
\usepackage{mathtools}
\usepackage{xcolor}

\usepackage{algcompatible,amsmath}

\usepackage{xcolor}
\usepackage[linesnumbered,ruled,vlined]{algorithm2e}

\SetCommentSty{mycommfont}

\usepackage{array}
\usepackage{tabularx,ragged2e,booktabs}
\usepackage{subcaption}

\setcopyright{rightsretained}

\acmDOI{}

\acmISBN{978-3-89318-083-7}

\acmConference[EDBT 2020]{22nd International Conference on Extending Database Technology (EDBT)}{March 30-April 2, 2020}{Copenhagen, Denmark} 
\acmYear{2020}

\begin{document}
\title{Ensemble Grammar Induction For Detecting Anomalies in Time Series}

\author{Yifeng Gao}
\affiliation{%
  \institution{George Mason University}
  \streetaddress{4400 University Drive}
  \city{Fairfax} 
  \state{VA} 
  \postcode{22030}
}
\email{ygao12@gmu.edu}

\author{Jessica Lin}
\affiliation{%
  \institution{George Mason University}
  \streetaddress{4400 University Drive}
  \city{Fairfax} 
  \state{VA} 
  \postcode{22030}
}
\email{jessica@gmu.edu}

\author{Constantin Brif}
\affiliation{%
  \institution{Sandia National Laboratories}
  \streetaddress{CA 94550, USA}
  \city{Livermore} 
  \state{CA}}
\email{cnbrif@sandia.gov}

\renewcommand{\shortauthors}{}

\begin{abstract}

Time series anomaly detection is an important task, with applications in a broad variety of domains. Many approaches have been proposed in recent years, but often they require that the length of the anomalies be known in advance and provided as an input parameter. This limits the practicality of the algorithms, as such information is often unknown in advance, or anomalies with different lengths might co-exist in the data. To address this limitation, previously, a linear time anomaly detection algorithm based on grammar induction has been proposed. While the algorithm can find variable-length patterns, it still requires preselecting values for at least two parameters at the discretization step. How to choose these parameter values properly is still an open problem. In this paper, we introduce a grammar-induction-based anomaly detection method utilizing ensemble learning. Instead of using a particular choice of parameter values for anomaly detection, the method generates the final result based on a set of results obtained using different parameter values. We demonstrate that the proposed ensemble approach can outperform existing grammar-induction-based approaches with different criteria for selection of parameter values. We also show that the proposed approach can achieve  performance similar to that of the state-of-the-art distance-based anomaly detection. 
\end{abstract}

\maketitle

\section{Introduction}

Time series anomaly detection is an important task, with applications in a broad variety of domains. Many approaches have been proposed in recent years, but often they require that the length of the anomalies be known in advance and provided as an input parameter. Recently, this limitation has been addressed by introducing a time series anomaly detection approach that is based on grammar induction and has a linear time complexity with respect to the data size (the time series length).

Typically, the grammar-induction-based anomaly detection follows a four-step process. In the first step, the input time series is converted into a discrete sequence of symbols via a sliding window; this discretization depends on two parameters. In the second step, grammar induction (e.g., via the Sequitur algorithm \cite{nevill1997identifying}) is applied to the discrete sequence to quickly identify grammar rules that are repeating strings of symbols. The third step maps the repeating strings back to the time series subsequences that the strings represent. Finally, a meta time series named \emph{rule density curve} is computed, which records the frequency of grammar rules at each point and is used to detect and rank the anomalies. Specifically, it is assumed that anomalies correspond to rarely occurring strings and hence are indicated by minima of the rule density curve.

While the grammar-induction-based algorithm can find variable-length patterns, it still requires preselecting values for at least two parameters at the discretization step. These two discretization parameters are the number of segments (i.e., the length of a word, also called PAA size which will be explained later), and the alphabet size. How to choose the parameter values properly is still an open problem, especially in an unsupervised setting where no training data are available.

\begin{figure}[ht]
 \centering
 \includegraphics[width=80mm]{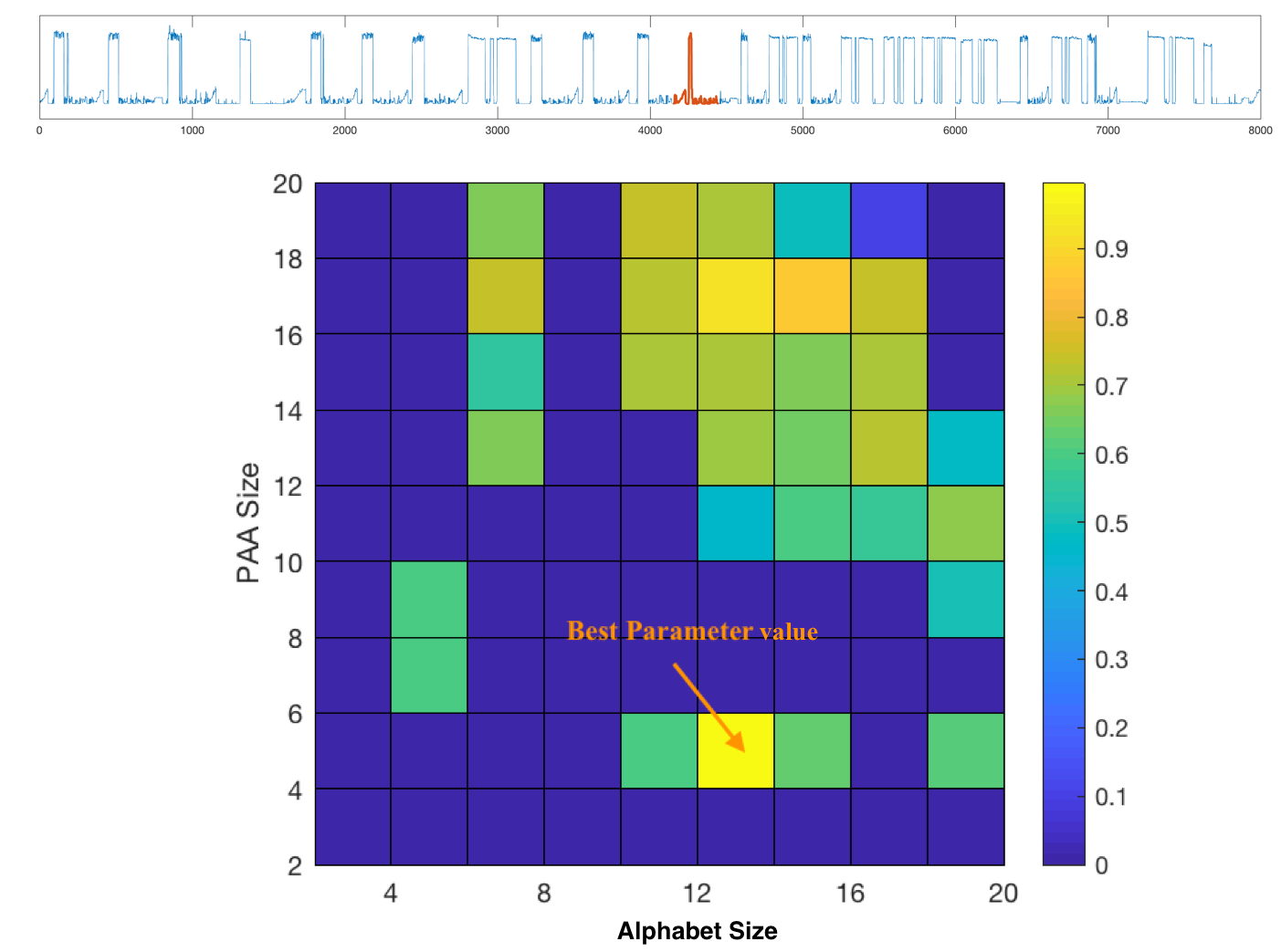}
 \caption{(top) A snippet of dishwasher electricity usage time series. An anomalous cycle is highlighted in red. (bottom) Performance of the grammar-induction-based algorithm with different discretization parameter values in the task of detecting the anomalous subsequence.}
 \label{fig:demo}
\end{figure}

Figure \ref{fig:demo} presents an example that illustrates the challenge of choosing proper parameter values, even with the knowledge of ground truth information that allows us to evaluate the quality of anomalies detected. Figure \ref{fig:demo}.top shows a snippet of dishwasher electricity usage time series. An anomalous cycle that has an unusual short power usage period is highlighted in red. We ran the algorithm with different parameter value combinations, and the results are shown in Figure \ref{fig:demo}.bottom. According to the figure, the best parameter value combination (indicated by arrow) is significantly different from the second best combination. Moreover, we see that parameter values that are close to the optimal values actually perform badly. As a result, guessing the ``best'' parameter values can be very tricky.  

To overcome the challenge mentioned above, in this paper, we introduce a robust version of the grammar-induction-based anomaly detection method, which utilizes ensemble learning. Intuitively, instead of using a single combination of parameter values, the method generates the final result based on a set of results obtained using different parameter values. We design an approach that combines the results returned by each ensemble member corresponding to a different combination of parameter values. Furthermore, in the discretization step, we use an approach to compute multi-resolution words to efficiently represent time series subsequences. This approach can dramatically reduce the cost of discretization and increase the scalability of the proposed work.

The contribution of this paper is summarized as follows:

\begin{itemize}
    \item The proposed ensemble grammar induction can achieve better performance than the grammar induction with a single combination of parameter values. 
    \item The proposed ensemble grammar induction is particularly suitable for unsupervised anomaly detection, where no training set is available to perform a grid search for the best parameter values.
    \item The proposed approach has a linear time complexity with respect to the data size and can achieve performance comparable to that of the state-of-the-art distance-based approach that has a quadratic time complexity.
    \item We adapt a fast algorithm to discretize subsequences, which reduces the cost of repeated computation.
    \item We demonstrate that the ensemble approach can be applied to find meaningful anomalies in real-world application.
\end{itemize}

The rest of the paper is organized as follows. Section 2 describes related work. Section 3 presents definitions and notations used in the paper. The process used for discretization and numerosity reduction is described in Section 4. The grammar-induction-based anomaly detection is presented in Section 5. The ensemble approach is introduced in Section 6. The experimental results are shown in Section 7, and conclusions are summarized in Section 8.

\section{Related Work}

First, we describe the state-of-the-art time series anomaly detection approach. Keogh et al. \cite{keogh2005hot} introduced a concept named \emph{time series discord}, which is the subsequence that has the largest one--nearest-neighbor (1-NN) distance, hence it represents the most unusual subsequence in the time series. The authors introduced an algorithm named HOTSAX to effectively detect the time series discord. Recently, a series of matrix-profile-based approaches, STOMP \cite{zhumatrix} and STAMP \cite{yeh2016matrix}, have been introduced for fast computation of 1-NN distances for every subsequence. It has been shown that compared with the original method, STOMP and STAMP can achieve more stable and generally better performance. However, all these methods have a quadratic time complexity with respect to the data size, and the accuracy is sensitive to the  length of the discord that has to be specified in advance as an input parameter \cite{senin2014grammarviz}.

In previous work, we proposed a series of approximate time series pattern discovery algorithms called GrammarViz based on grammar induction \cite{li2012visualizing,senin2014grammarviz,senin2018grammarviz}. The idea is that by learning a context-free grammar from a discrete sequence of symbols that approximates the original time series, one can identify the repeating strings. Recently, in \cite{senin2015time}, we have extended the idea of grammar-induction-based anomaly detection by introducing the concept of rule density curve, which is a meta time series that records the frequency of grammar rules (repeating strings) at each point of the original time series. Since anomalies correspond to rarely occurring strings, they are indicated by minima of the rule density curve. It has been shown that this approach can achieve competitive performance compared to the state-of-the-art while having linear time complexity. However, the algorithm requires the user to preselect values of two important parameters, and the performance can be greatly affected by these values.

Several ensemble algorithms have been proposed for unsupervised anomaly detection \cite{rayana2015less,zhao2019lscp}. Most of them use the average performance as the ground truth and select a very small number of detectors to approximate the average performance. All these approaches are introduced for point-based anomaly detection. How to use them for detecting anomalous subsequences is still unclear. Besides, since subsequences extracted from a time series via a sliding window are highly overlapped, the detector used in these approaches \cite{breunig2000lof} cannot resolve this problem. 

Other works \cite{chen2008multi,hemalatha2015minimal,feremans2019pattern} also focus on anomaly detection; however, these techniques often focus on detecting anomalies in event logs (discrete or mixed data type time series) and the lengths of the anomalies are often very short. 

\section{Notations and Problem Definition}

We first describe the fundamental definitions related to time series and grammar induction. We then formulate the problem of time series anomaly detection.

\subsection{Notations and Definitions}

We start with the definitions related to time series:

\textbf{Time series} $T=t_1,\dots,t_N$ is a set of observations ordered by time. 

\textbf{Subsequence} $T_{p,q}$ of a time series $T$ is a subsequence of elements in $T$ starting from position $p$ and ending at position $q$, of length $n=q-p+1$. Typically, $n \ll N$,  and $1 \leq p \leq N-n+1$.

Subsequences can be extracted from time series via a sliding window. In many applications, we are interested in finding unusual ``shapes.'' Therefore, anomaly discovery result is more meaningful when the method can maintain offset- and amplitude-invariance during the anomaly detection process. This can be achieved by normalizing all subsequences prior to applying an anomaly detection algorithm. \textbf{z-normalization} is a procedure that normalizes the mean and standard deviation of a subsequence to zero and one, respectively.

\begin{figure}[ht]
 \centering
 \includegraphics[width=70mm]{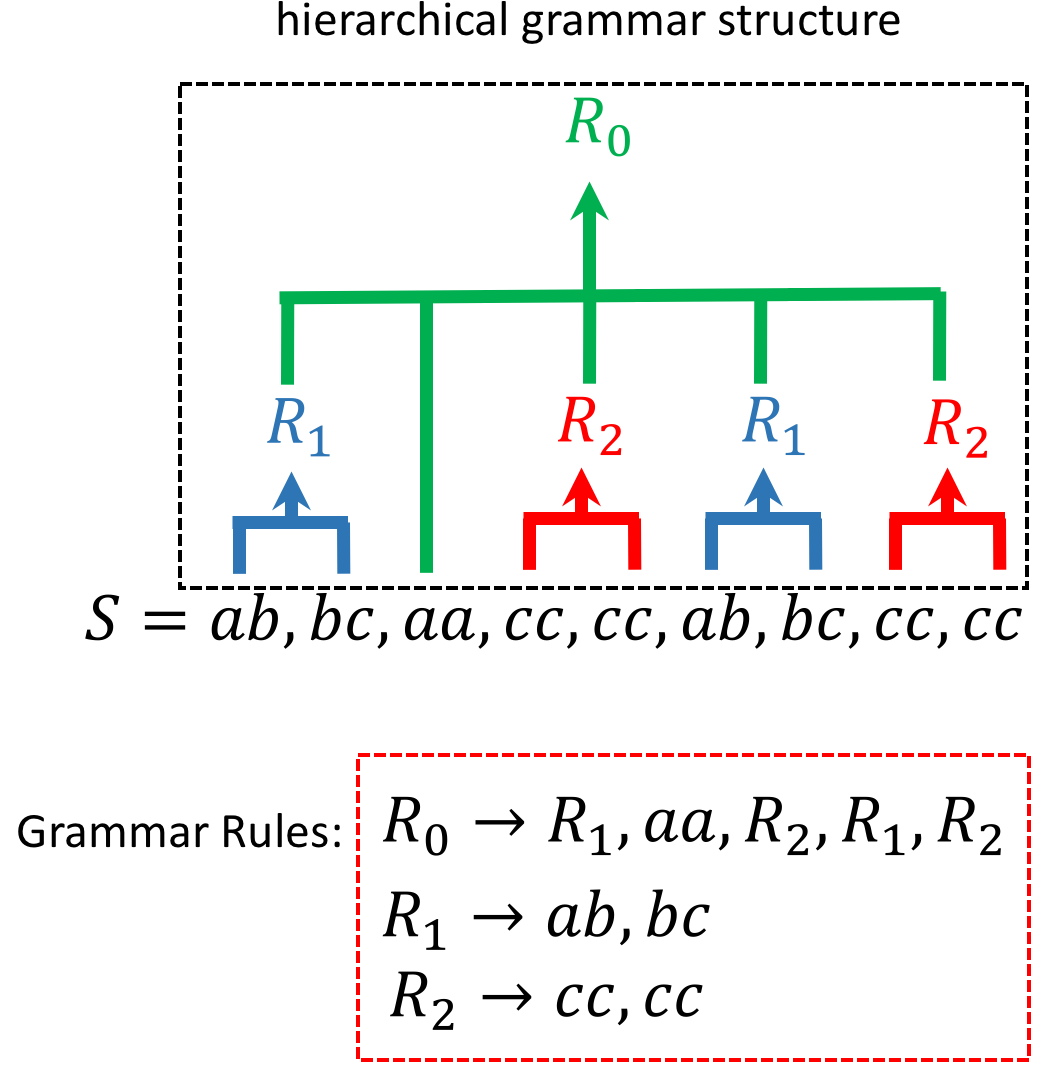}
 \caption{Example of grammar induction applied to a token sequence.}
 \label{fig:rule}
\end{figure}

Since the proposed anomaly detection approach is based on grammar induction, we next introduce the definitions related to grammar induction using a toy example shown in Figure \ref{fig:rule}. Intuitively, grammar induction is a process that induces a hierarchical grammar structure from a \textbf{token sequence} $S$, which is a sequence of discrete tokens (words). In the example, $S$ consists of nine two-letter tokens. The hierarchical grammar structure is represented by a set of \textbf{grammar rules}. Each grammar rule represents a repeating string of token segments in $S$. In the example, two rules $R_1$ and $R_2$ represent the repeating token segments $ab,bc$ and $cc,cc$, respectively. Following the terminology used in previous work \cite{nevill1997identifying,senin2014grammarviz}, each grammar rule is also called a \textbf{non-terminal} and each token stored in the token sequence is called a \textbf{terminal}.

Using the set of grammar rules, the original token sequence $S$ is represented by the compressed sequence $R_0$. Since compressibility is a measure of regularity (and hence incompressibility is a measure of anomalousness), the compressed sequence plays an important role in motif (repeated patterns) discovery and anomaly detection.

Clearly, the grammar induction approach cannot be directly used for real-valued time series. To use it, we need first to approximate the original time series by a discretized token sequence. We describe the time series discretization process in detail in Section 4.

\subsection{Problem of Anomaly Detection}

Based on the definitions above, we can formulate the problem of anomaly detection. In this work, we follow the previous anomaly detection framework GrammarViz \cite{senin2014grammarviz}, and determine a hierarchical grammar structure for a time series, through the processes of time series discretization and grammar induction. The anomaly candidates are subsequences that cannot be compressed by induced grammar rules. 

To illustrate the basic idea behind this process of anomaly detection, we use another toy example of a simple token sequence:
\begin{equation}
\label{eq:S-example}
    S=aa,bb,cc,xx,aa,bb,cc .
\end{equation}
The grammar structure induced from $S$ is shown in Table \ref{tab:gi}. We see that $S$ contains a repeating pattern, $aa,bb,cc$, represented by the grammar rule R1. The token $xx$, however, does not appear in any grammar rules (i.e., it is incompressible). It is not hard to see that token $xx$ is \textit{structurally} dissimilar from the rest of the sequence. In the grammar-induction-based time series anomaly detection framework, the subsequence that $xx$ represents is considered an anomaly candidate. 

\begin{table}[ht]
    \centering
    \caption{Example of grammar rules induced from token sequence $S$ of Eq.~\eqref{eq:S-example}}

\begin{tabular}{|c|l|}

  \hline
   Grammar rules  & Expanded sequence\\
  \hline\hline
   $R_0 \rightarrow R_1, xx, R_1$ & $aa,bb,cc,xx,aa,bb,cc$ \\
  \hline
   $R_1 \rightarrow aa,bb,cc$ &  $aa,bb,cc$\\

  \hline
\end{tabular}
    \label{tab:gi}
\end{table}

\section{Discretization and Numerosity Reduction}

Time series discretization \cite{lin2007experiencing} is a common step in many time series data mining tasks, including anomaly detection \cite{keogh2005hot,gao2017trajviz}. Since grammar induction requires discrete input, it is necessary to approximate the time series by a token sequence first. In general, discretization offers several advantages including noise removal, dimensionality reduction, and improved efficiency. 

In this section, we first describe the algorithm called \emph{Symbolic Aggregate approXimation} (SAX), a widely used time series discretization technique. We then describe \emph{numerosity reduction}, a procedure that removes repeating consecutive tokens to form a more compact token representation of a time series \cite{lin2004visually,senin2014grammarviz}.

\subsection{Symbolic Aggregate approXimation }
  
In this section, we describe SAX, a popular technique used to discretize univariate time series. 

SAX consists of two steps. At the first step, the \emph{Piecewise Aggregate Approximation} (PAA) is used to convert a normalized subsequence of length $n$ from a time series $T$ into a representation of a lower dimension $w < n$ ($w$ is called PAA size). Specifically, the subsequence is divided into $w$ equal-sized windows, and the average value of the elements within each window is computed. In other words, the PAA coefficients vector \cite{lin2007experiencing} is a $w$-dimensional vector that consists of the average values from $w$ equal-sized segments of the input subsequence. PAA coefficients are an approximate representation of the original subsequence. 

At the second step, the PAA coefficients vector is mapped to $w$ symbols from an alphabet of size $a$, according to a breakpoint table \cite{lin2007experiencing}, defined such that the regions are approximately equal-probable under the Gaussian distribution. This maximizes the chances that the symbols occur with an approximately equal probability. These $w$ symbols form a SAX word. 

\begin{figure}[ht]
 \centering
 \includegraphics[width=75mm]{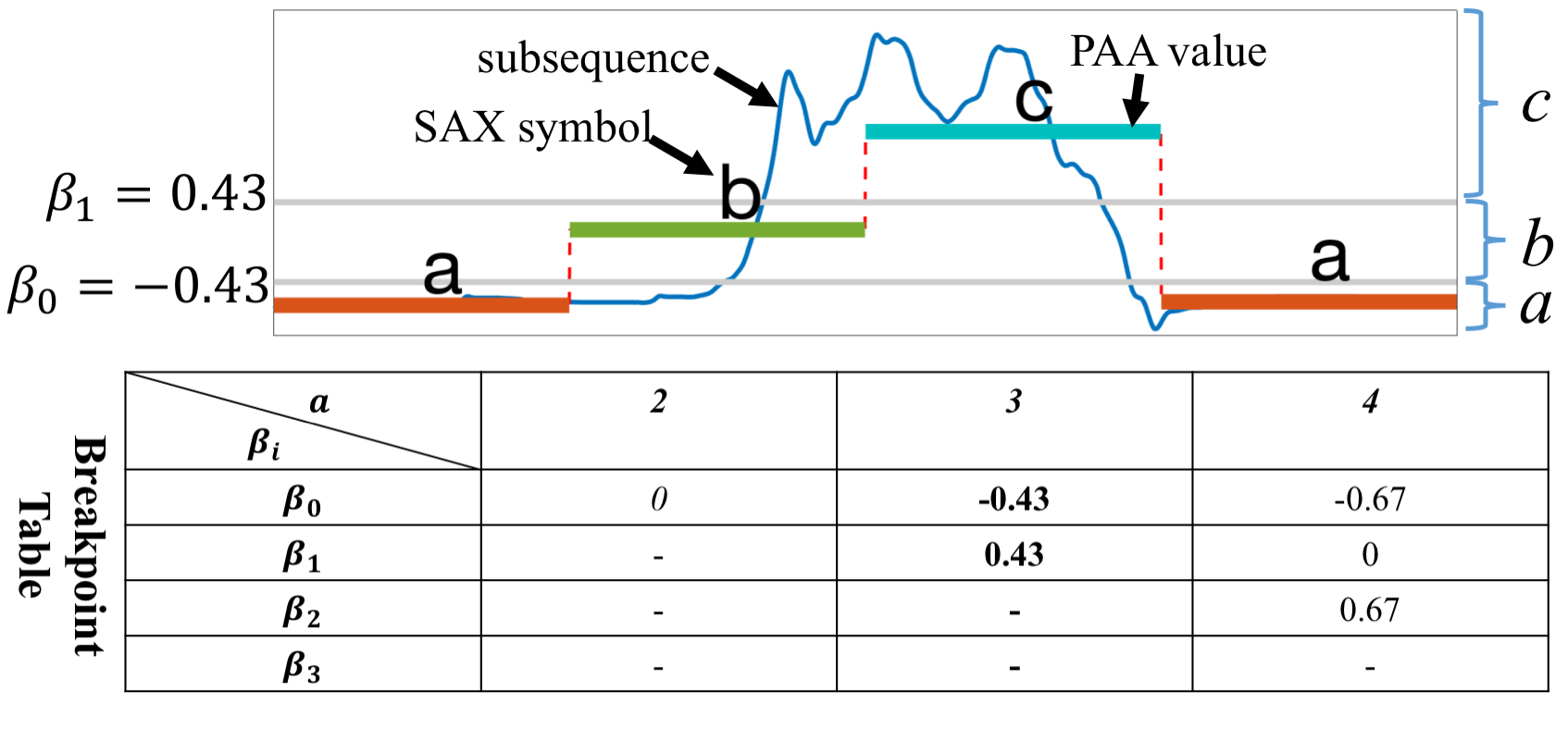}
 \caption{Example of the SAX process, with $a=3$ and $w=4$, which approximates the original subsequence by the word $abca$.}
 \label{fig:sax}
\end{figure}

Figure \ref{fig:sax} illustrates the SAX process for an example subsequence (shown as the blue curve). The bold flat lines represent the values of PAA coefficients computed from their respective segments in the subsequence. The breakpoint table with $a$ from 2 to 4 is also shown in the figure. Since we set $a=3$ in this example, two breakpoints in the second column of the table are used to generate three regions: $(-\infty,-0.43),[-0.43,0.43),[0.43,\infty)$. The PAA coefficients falling into these three regions are mapped to symbols $a$, $b$, and $c$, respectively. In this example, the SAX word $abca$ is formed to approximate the original subsequence.

This description shows how SAX is applied to a given subsequence. In order to discretize the entire time series, SAX is usually applied via a sliding window of length $n$.

\subsection{Numerosity Reduction}

In practice, since neighboring subsequences are only off by one point, the neighboring tokens generated by SAX are often identical to each other. This phenomenon, however, will result in an overwhelming number of grammar rules that represent trivial matches, and this redundancy will significantly affect both the scalability and the quality of results. To avoid this problem, a \emph{numerosity reduction} step is added to further compress the token sequence. Specifically, whenever there is a sequence of one token repeating consecutively multiple times, numerosity reduction will output only the first occurrence of the token along with its offset. 

For example, a token sequence 
\begin{equation}
    S = ba,ba,ba,dc,dc,aa,ac,ac
\end{equation}
can be compressed to 
\begin{equation}
S_{\mathrm{NR}} = ba_1,dc_4,aa_6,ac_7 ,
\end{equation}
where the subscripts indicate the positions in the original uncompressed token sequence. $S_{\mathrm{NR}}$ contains all information needed to retrieve the original token sequence.

\begin{table*}[t]
    \centering
    \caption{Example of Sequitur inducing grammar from token sequence $S_{\mathrm{NR}}$ of Eq.~\eqref{eq:S-NR-example}}
\scalebox{0.8}{
\begin{tabular}{|c|l|l|}
  \hline
  Step & Grammar rules & Digrams \\
  \hline\hline
  1. & $S\rightarrow ab_1$ & \\
  \hline
  2. & $S\rightarrow ab_1, bc_8$ & $\{ab,bc\}$ \\ 
  \hline
  3. & $S\rightarrow ab_1, bc_8, aa_{15}$ & $\{ab,bc\},\{bc,aa\}$\\ 
  \hline
  4. & $S\rightarrow ab_1, bc_8, aa_{15}, cc_{21}$ & $\{ab,bc\},\{bc,aa\},\{aa,cc\}$\\ 
  \hline
  5. & $S\rightarrow ab_1, bc_8, aa_{15}, cc_{21}, ca_{25}$ & $\{ab,bc\},\{bc,aa\},\{aa,cc\}, \{cc,ca\}$\\
  \hline
    6. & $S\rightarrow ab_1, bc_8, aa_{15}, cc_{21}, ca_{25}, ab_{29}$ & $\{ab,bc\},\{bc,aa\},\{aa,cc\}, \{cc,ca\}, \{ca,ab\}$ \\
  \hline
  7. & $S\rightarrow \boldsymbol{ab_1, bc_8}, aa_{15}, cc_{21}, ca_{25}, \boldsymbol{ab_{29}, bc_{34}}$ & $\boldsymbol{\{ab,bc\}},\{bc,aa\},\{aa,cc\}, \{cc,ca\}, \{ca,ab\}, \boldsymbol{\{ab,bc\}}$\\
  \hline
  8. & $S\rightarrow R_1, aa_{15}, cc_{21}, ca_{25}, R_1$  & $\{R_1,aa\},\{bc,aa\},\{aa,cc\}, \{cc,ca\}, \{ca,ab\},\{ca,R_1\}$\\
  &$R_1 \rightarrow ab, bc$ & $\{ab,bc\}$ \\
  \hline
  9. & $S\rightarrow \boldsymbol{R_1, aa_{15}}, cc_{21}, ca_{25},\boldsymbol{R_1, aa_{40}}$ & $\boldsymbol{\{R_1,aa\}},\{bc,aa\},\{aa,cc\}, \{cc,ca\}, \{ca,ab\},\{ca,R_1\},\boldsymbol{\{R_1,aa\}}$\\
  
  &$R_1 \rightarrow ab, bc$ & $\{ab,bc\}$ \\
  \hline
  10. & $S\rightarrow R_2, cc_{21}, ca_{25}, R_2$ & $\{R_2,cc\},\{cc,ca\},\{ca,R_2\}$\\
  &$\boldsymbol{R_1 \rightarrow ab, bc}$ & $\{ab,bc\}$ \\
  &$R_2 \rightarrow R_1, aa$ & $\boldsymbol{\{R1,aa\}}$ \\
  \hline
  11. & $S\rightarrow R_2, cc_{21}, ca_{25}, R_2$ & $\{R_2,cc_{21}\},\{cc,ca\},\{cc_{21},R_2\}$\\
  &$R_2 \rightarrow ab,bc,aa$ & $\{ab,bc\},\{bc,aa\}$\\
  \hline
\end{tabular}
}
    \label{tab:seq}
\end{table*}

\section{Grammar Induction}

In this section, we first describe Sequitur \cite{nevill1997identifying}, a grammar induction algorithm with a linear time complexity. We then describe the process of computing the rule density curve and ranking anomalies.

\subsection{Sequitur}

Sequitur is a linear-time greedy algorithm to induce a context-free grammar structure from a discrete token sequence. 

Sequitur maintains a list of grammar rules and a table of digrams  based on the input sequence. A digram is a pair of consecutive tokens (terminals or non-terminals) in the sequence. Two principles, digram uniqueness and rule utility, are applied to constrain the rules during grammar induction. Digram uniqueness requires that digrams stored in the digram table should be unique. Rule utility requires that rules that only appear once should be removed to minimize the size of grammar.

To illustrate how Sequitur works, consider an example token sequence generated by SAX with parameters $w=2$, $a=3$, $n=16$, after the numerosity reduction step: 
\begin{equation}
\label{eq:S-NR-example}
S_{\mathrm{NR}} = ab_{1}, bc_{8}, aa_{15}, cc_{21}, ca_{25}, ab_{29}, bc_{34}, aa_{40} .
\end{equation}
A step-by-step grammar induction process is shown in Table \ref{tab:seq}. From Step 1 to Step 6, since neither digram uniqueness nor rule utility is violated, the algorithm simply reads the first six tokens and adds digrams into the digram table, respectively. In Step 7, the algorithm finds that the digram $\{ab,bc\}$ occurs in the digram table twice. Therefore, in Step 8, the algorithm forms a new rule, $R_1 \rightarrow ab, bc$. The new non-terminal symbol $R_1$ is generated to replace all occurrences of $\{ab,bc\}$, and the digram table is updated to maintain the uniqueness of digrams by removing $\{ab,bc\}$ and adding $\{R_1,aa\}$ and $\{ca,R_1\}$. Similarly, in Step 9, the digram $\{R_1,aa\}$ appears twice and therefore is replaced, in Step 10, by a new rule $R_2$. After the digram table is updated in Step 10, the algorithm finds that the rule $R_1$ only appears once. Therefore, in Step 11, $R_1$ is expanded to satisfy rule utility. 

After processing the entire token sequence, the original token sequence is compressed into $R_0 \rightarrow R_2, cc, ca, R_2$ and the string $cc,ca$ is identified as an anomaly candidate since it cannot be compressed.

\subsection{Rule Density Curve}
\label{sec:RDC}

We next describe the construction of the rule density curve \cite{senin2014grammarviz}. 
Simply stated, a rule density curve is a meta time series, in which each value is equal to the number of grammar rules that ``cover'' the respective time point. For anomaly detection, we are interested in the intervals for which the rule densities are the lowest. These intervals correspond to subsequences (more precisely, the SAX strings that represent these subsequences) that rarely appear in grammar rules, and hence are potentially anomalous. For the example sequence shown in Table \ref{tab:seq}, the time series points corresponding to the token subsequence $cc,ca$ will have a count of zero since $cc,ca$ does not appear in any grammar rules.

To construct the rule density curve, we map each instance of a rule back to the subsequence index based on the index recorded in the numerosity reduction step, and then we keep track of the number of rules that cover each time point.

\begin{figure}[ht]
 \centering
 \includegraphics[width=80mm]{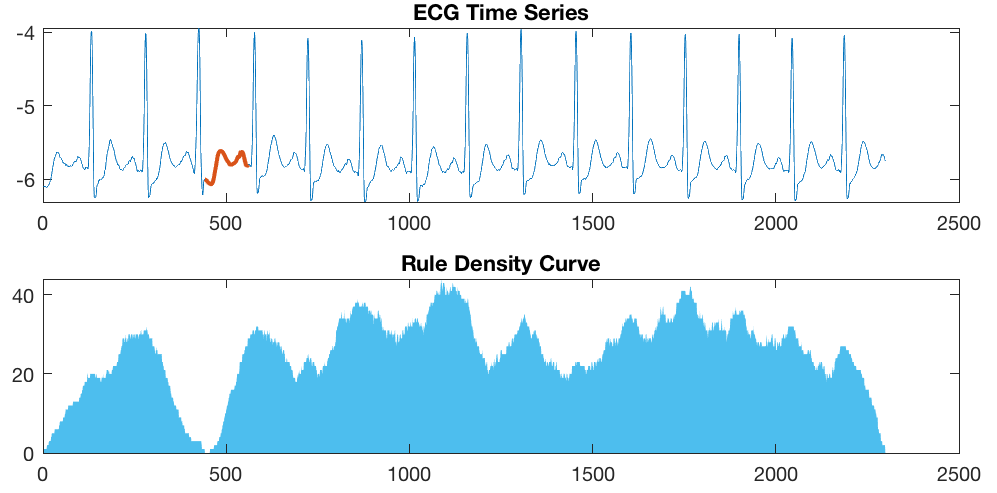}
 \caption{An example of the rule density curve generated for an ECG time series.}
 \label{fig:curve}
\end{figure}

Once the rule density curve is constructed, we can locate the potentially anomalous subsequences by finding the local minima of the curve and ranking them based on their respective rule density values.

An example is shown in Figure \ref{fig:curve}. Figure \ref{fig:curve}.top shows an electrocardiogram (ECG) time series and Figure \ref{fig:curve}.bottom shows the rule density curve computed from this time series. An anomaly candidate, highlighted in red in Figure \ref{fig:curve}.top, corresponds to the minimum of the rule density curve. According to \cite{senin2014grammarviz}, this location corresponds to an anomalous premature heart beat.

\subsection{Challenges of Parameter Selection}

Although the grammar-induction-based approach described above has been successfully used in time series data mining problems \cite{senin2014grammarviz,itr}, it has two drawbacks that can strongly affect the quality of patterns found. First, approximation errors are inevitably introduced due to the information loss at the discretization step. For example, two subsequences represented by the same SAX word may actually be dissimilar and have a large distance. Second, Sequitur is a greedy algorithm which cannot guarantee the globally optimal result. Therefore, the grammar rules learned may not perfectly reflect the repeating and anomalous patterns in the time series. In summary, a single run of grammar induction with fixed parameter values may simply not be enough to detect high quality anomalies. 

In this paper, in order to mitigate the limitations of the existing grammar-induction-based anomaly detection framework while still maintaining high efficiency, we introduce an ensemble approach to generate a rule density curve based on multiple runs with different parameter values. 

\section{Ensemble Grammar Induction}

In this section, we introduce the proposed ensemble-based grammar induction approach.

\subsection{Ensemble Rule Density Curve}

The algorithm that generates the ensemble rule density curve from a set of multiple grammar induction runs is shown in Algorithm 1. First, we generate $N$ rule density curves using values for PAA size $w$ and alphabet size $a$ randomly chosen from intervals $[2,w_{\mathrm{max}}]$ and $[2,a_{\mathrm{max}}]$, respectively (Lines 4--6). Second, we remove low-quality curves based on their standard deviations (Lines 7--10). Third, we normalize each remaining curve to the same scale (Line 11). Finally, we combine the results from all normalized curves by computing the median (Line 14).

\begin{algorithm}[ht]
    \caption{Ensemble Rule Density Curve}
  \begin{algorithmic}[1]
    \STATE \textbf{Input}: time series $T$, sliding window length $n$, ensemble size~$N$, maximum PAA size $w_{\mathrm{max}}$, maximum alphabet size~$a_{\mathrm{max}}$, ensemble selectivity~$\tau$
    \STATE \textbf{Output}: ensemble rule density curve $d_{\mathrm{e}}$
    \STATE $\mathcal{D}=$[]
    \COMMENT{Compute $N$ rule density curves with random parameter values and compute the standard deviation for~each of them}
        
    \FOR{$i = 1$ to $N$}
      
      \COMMENT{Randomly generate parameter values; any $w,a$ combination is used only once}
      
      \STATE $w,a = \text{GenerateRandomParam}(w_{\mathrm{max}}, a_{\mathrm{max}})$
      
      \STATE $d_i = \text{GrammarInduction}(T,n,w,a)$
      \STATE $s_i = \text{ComputeStd}(d_i$);
    \ENDFOR
    \STATE index = ArgSort($s$)
    \COMMENT{Sort the standard deviations in descending order}
    \FOR{$i = 1$ to $\tau N$}
    \COMMENT{Keep $\tau\%$ of the density curves and normalize each of them}
    \STATE $d_{\mathrm{norm}} = d_{\mathrm{index}[i]}/\mathrm{max}(d_{\mathrm{index}[i]})$
    \STATE $\mathcal{D}$.add($d_{\mathrm{norm}}$)
    \ENDFOR
    \STATE $d_{\mathrm{e}} = \text{ComputeMedian}(\mathcal{D})$
    \COMMENT{Compute the median for all normalized rule density curves}
    \STATE \textbf{return} $d_{\mathrm{e}}$
  \end{algorithmic}
\end{algorithm}

\subsubsection{Removing Low-Quality Rule Density Curves}

Not all rule density curves provide reliable information about anomalies. Intuitively, low-quality curves correspond to situations where a grammar rule set has a similar frequency everywhere, and they should be removed from the ensemble to improve the effectiveness of anomaly detection. Towards this end, the algorithm computes the standard deviation for each of the generated rule density curves (Line 8). We then rank the curves based on the standard deviation in descending order and only keep the top $\tau\%$ of the curves to form the ensemble set $\mathcal{D}$.  

\begin{figure}[ht]
 \centering
 \includegraphics[width=75mm]{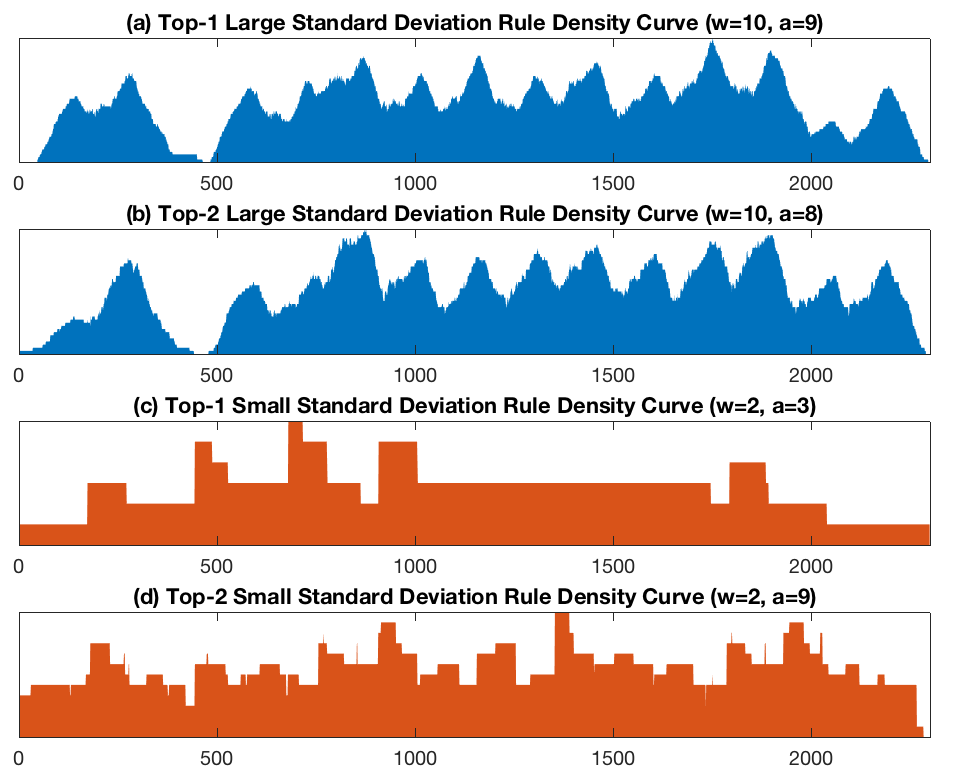}
 \caption{Examples of rule density curves generated from the ECG time series with different parameter values.}
 \label{fig:bad}
\end{figure}

An example that illustrates the quality variation among different rule density curves is shown in Figure \ref{fig:bad}. All four curves in Figure \ref{fig:bad} are generated from the ECG time series shown in Figure \ref{fig:curve}.top. The first two curves colored in blue are the top-2 curves based on the ranking by the standard deviation. The last two curves colored in red are the bottom-2 curves based on this ranking. As seen from the figure, it is very hard to determine the anomaly location from the bottom-2 curves. In contrast, the top-2 curves reveal the anomaly location clearly. This example demonstrates that using the rule density curves with higher standard deviations can help identify anomalies and potentially reduce the number of false positives.

\subsubsection{Normalizing Rule Density Curves}

Intuitively, each rule density curve may have a different scale. For example, using a coarse discretization resolution (i.e., small PAA size and alphabet size) tends to result in a large average frequency of grammar rules since it is more likely to find a match due to a smaller dictionary size. Conversely, using a fine discretization resolution (i.e., large PAA size and alphabet size) tends to result in a small average frequency of grammar rules. Consequently, without a normalization process, some anomaly detectors may undeservedly dominate the decision. 

To avoid this problem, we normalize each of the rule density curves in the ensemble set, so that each point of the curve falls in the range [0, 1]. It is worth noting that we do not use min-max normalization because we want to preserve the significance of the locations where the rule density is zero.

\subsubsection{Combining Rule Density Curves}

At the final step, we compute an ensemble rule density curve $d_\mathrm{e}$ based on all normalized curves kept in the set. In this paper, the specific way in which we combine all rule density curves in the ensemble is by computing the median value at each time point. Once the ensemble rule density curve is generated, the algorithm ranks the anomaly candidates through the same process as described in Sec.~\ref{sec:RDC}.

\subsection{Multi-resolution SAX Word Computation}

Since the ensemble approach requires computing different SAX words (corresponding to different parameter values) for the same subsequence, it is important to increase the efficiency of the discretization process.  Therefore, in this subsection, we describe a fast way to compute multi-resolution SAX words \cite{gao2018exploring}. 

\subsubsection{Fast Computation of SAX Words with Different Values of $w$}

We first describe a fast way to compute the PAA coefficients \cite{rakthanmanon2012searching}. First, two vectors of statistical features for a time series $T$ are pre-computed: $\mathrm{ESum}_x(x)=\sum_{i=1}^{x}t_{i}$ and $\mathrm{ESum}_{xx}(x)=\sum_{i=1}^{x}t_{i}^2$. Given a subsequence $T_{p,q}$ of length $n$, the SAX representation can be computed by Algorithm 2. In this algorithm, the mean and variance of $T_{p,q}$ are computed in constant time (Lines 3--5). The cost of computing the PAA coefficients (Lines 6--8) is $O(w)$ for a single resolution, which is faster than the trivial approach whose computation cost is $O(n)$ ($w < n$). 

\begin{algorithm}[ht]
    \caption{Fast Compute PAA (FastPAA)}
  \begin{algorithmic}[1]
    \STATE \textbf{Input}: subsequence $T_{p,q}$, $\mathrm{Esum}_x$,$\mathrm{Esum}_{xx}$, PAA size $w$
    \STATE \textbf{Output}: PAA representation $A$
    \STATE $E_x = \mathrm{Esum}_x(q) - \mathrm{Esum}_x(p)$ 
    \STATE $E_{xx} = \mathrm{Esum}_{xx}(q) - \mathrm{Esum}_{xx}(p)$
    \STATE $n = q-p+1$,  $\mu = E_{x}/n$, $\sigma = \sqrt{(E_{xx}-E_{x}^2/n)/(n-1)}$
    \FOR{every PAA segment}
     \STATE $A_i = \left( \frac{\mathrm{Esum}_{x}(A_{i,\mathrm{e}}) - \mathrm{Esum}_{x}(A_{i,\mathrm{s}})}{n/w} - \mu \right)/\sigma$
     \COMMENT{$A_{i,\mathrm{s}}$ and $A_{i,\mathrm{e}}$ are the start and end points of the $i$th PAA segment}
     \ENDFOR
     \STATE \textbf{return} $A$
  \end{algorithmic}
\end{algorithm}

\subsubsection{Fast Computation of SAX Words with Different Values of $a$}

\begin{figure}[ht]
 \centering
 \includegraphics[width=78mm]{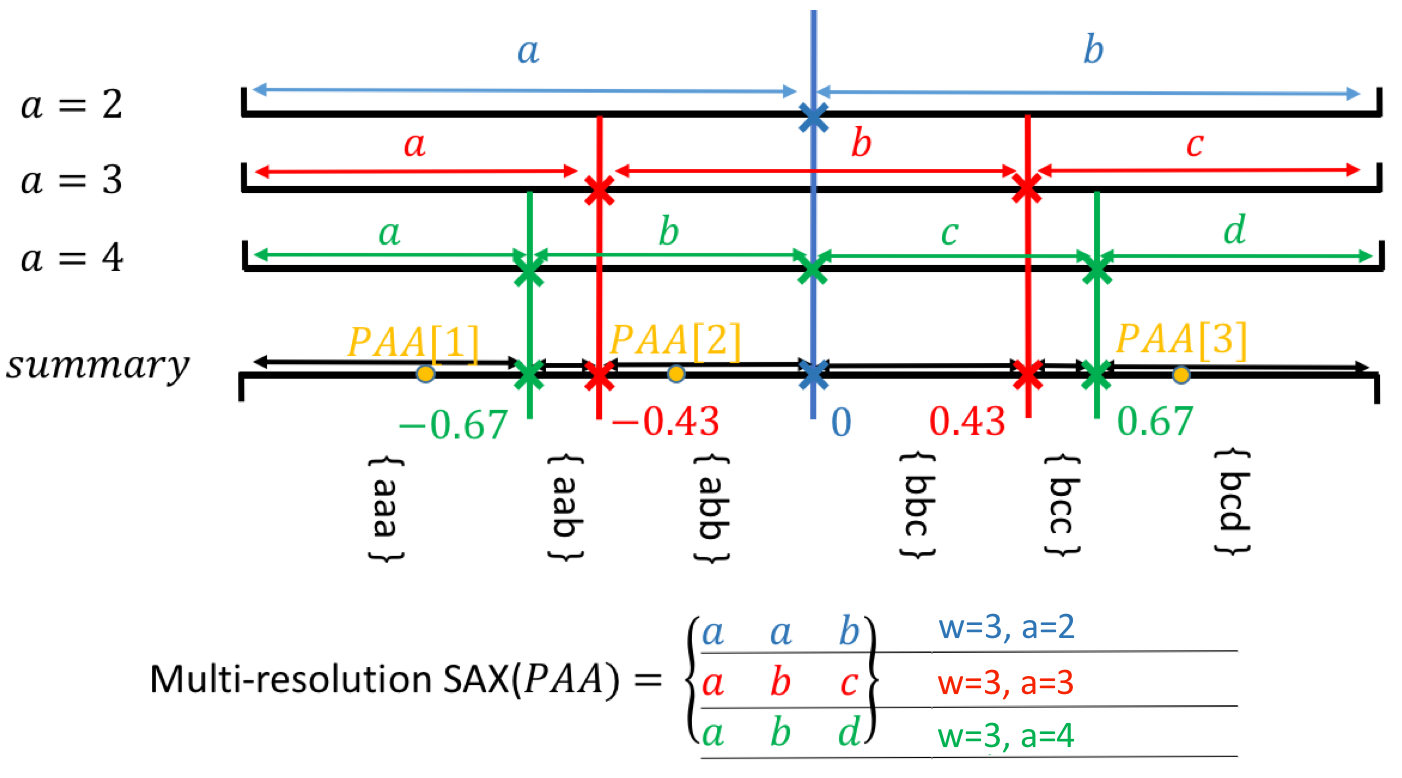}
 \caption{Fast computation of multi-resolution SAX.}
 \label{fig:msax}
\end{figure}

To efficiently compute SAX words with multiple resolutions, we adapt an algorithm we introduced in a previous work \cite{gao2018exploring}. Specifically, given a maximum alphabet size $a_{\mathrm{max}}$, to fast compute SAX words with different alphabet sizes, we first gather breakpoint tables of all alphabet size values used in the ensemble. For each interval between any two breakpoints, a symbol sequence containing corresponding symbols up to $a_{\mathrm{max}}$ resolution is recorded. We represent any PAA coefficient belonging to the interval by the pre-computed symbol sequence. 

An example with alphabet sizes from 2 to 4 is shown in Figure \ref{fig:msax}. In the figure, the set of breakpoints for all alphabet sizes from 2 to 4 (labeled by `$\times$' in each line) are projected to the line denoted as ``summary.'' All distinct breakpoints with $a$ from 2 to 4 in the breakpoint table (Fig.  \ref{fig:sax}.bottom) create 6 intervals, and each interval stores a sequence of symbols. The $i$th position in such a sequence stores the corresponding symbol for alphabet size $a=i+1$. For example, the 3 PAA coefficients that fall in intervals $(-\infty, -0.63]$, $(-0.43, 0]$ and $(0.63, \infty)$ (denoted by yellow dots) are mapped to symbol sequences $aaa$, $abb$, and $bcd$, respectively. A symbol matrix is then created by concatenating such symbol sequences (bottom of Figure \ref{fig:msax}). The $i$th row of the symbol matrix represents a SAX word generated with alphabet size $i+1$. For example, the first row $aab$ is the SAX word generated for $a=2$. To construct the matrix of SAX words with all alphabet sizes, for each PAA coefficient, we only need to perform at most 3 comparisons to determine the interval via binary search, which is the same as generating fixed-resolution SAX. By using binary search to determine which interval the PAA coefficient belongs to, we can find its SAX representations in all resolutions from $a = 2$ to $a = a_{\mathrm{max}}$ with a time complexity of $O(2 \log(a_{\mathrm{max}}))$. When $a_{\mathrm{max}} = 20$, the cost of computing all resolutions is similar to computing a fixed resolution.  

\subsubsection{Overall Improvement}
The time complexity of computing SAX words with multiple resolutions in a straightforward manner (without acceleration) is $O(nw_{\mathrm{max}}a_{\mathrm{max}}+w_{\mathrm{max}}^2a_{\mathrm{max}}^2)$. In contrast, with the proposed acceleration, the computation cost is $O(w_{\mathrm{max}}^2 \log(a_{\mathrm{max}}))$, which is a significant improvement.

\section{Experimental Evaluation}

We perform a series of numerical experiments to evaluate the accuracy and speed of ensemble grammar induction applied to time series anomaly detection. In all experiments, unless noted otherwise, the parameter values are: $a_{\mathrm{max}} = 10$, $w_{\mathrm{max}} = 10$, $N = 50$, and $\tau = 40\%$. All the experiments are conducted on a 16~GB~RAM laptop with quad core processor of 2.5~GHz. We first show the performance on real-world time series. We then evaluate the impact of parameter sets used in the ensemble. Finally, we analyze the scalability of the algorithm and conduct a case study on an electric load time series. 

\begin{table}[t]
\caption{Properties of datasets used for experimental evaluation}
\centering
\scalebox{0.8}{
\begin{tabular}{ |c|c|c|c|c| }
  \hline
  Dataset & Time Series  & Segment & Data Type \\
  & Length& Length & \\
    \hline
  TwoLeadECG &  1772 & 82 & ECG\\
  ECGFiveDay &  2772 & 132 & ECG\\
  GunPoint &  3150 & 150 & Motion\\
  Wafer &  3150 & 150 & Sensor\\
  Trace &  5775 & 275 & Sensor\\
  StarLightCurve &  21504 & 1024 & Sensor\\
   \hline
\end{tabular}
}
\label{tab:data}
\end{table}

\subsection{Performance Evaluation in Comparison to Baseline Methods}

Since it is difficult to find annotated time series for anomaly detection, to evaluate our proposed method, we compile our own data using several real-world and synthetic time series datasets from the UCR Time Series Classification Archive \cite{UCRArchive2018}. 

\subsubsection{Datasets}

\begin{table}[t]
\caption{Performance evaluation results (average Score)}
\centering
\scalebox{0.8}{
\begin{tabular}{ |c|c|c|c|c|c| }
  \hline
  Dataset  & Proposed   & GI-Random & GI-Fix & GI-Select & Discord  \\
  &  Approach  &  &  &  &  \\
    \hline\hline
  TwoLeadECG & 0.3951 &  0.2873& 0.0629 &  0.1663&  \textbf{0.4931} \\
  \hline
  ECGFiveDay &0.3903 &   0.2988&  0.2671 & 0.105& \textbf{0.4794} \\
    \hline
  GunPoint & \textbf{0.4728} &     0.3715&   0.2411 & 0.056& 0.4 \\
    \hline
  Wafer & \textbf{0.3179} & 0.2126 &   0.1382 & 0.248&  0.309\\
    \hline
  Trace & \textbf{0.5718} &   0.2022 &  0.3601&  0.3408&  0.2816\\
    \hline
  StarLightCurve & \textbf{0.9369}  &   0.6930&0.5301 & 0.8759 &   0.9161 \\
   \hline
\end{tabular}
}
\label{tab:score}
\end{table}

\begin{table}[t]
\caption{Performance evaluation results (HitRate)}
\centering
\scalebox{0.8}{
\begin{tabular}{ |c|c|c|c|c|c| }
  \hline
  Dataset  & Proposed  & GI-Random & GI-Fix & GI-Select & Discord  \\
    &  Approach  &  &  &  &  \\
    \hline\hline
  TwoLeadECG & 0.72 & 0.52 & 0.4 & 0.24 & 0.8\\
  \hline
  ECGFiveDay & 0.8 & 0.44 & 0.36 & 0.24 & 0.8\\
    \hline
  GunPoint & 0.68 & 0.56 & 0.44 & 0.12 & 0.68\\
    \hline
  Wafer & 0.72 & 0.4 & 0.36 & 0.4 & 0.52\\
    \hline
  Trace & 0.96 & 0.4 & 0.8 & 0.6 & 0.52\\
    \hline
  StarLightCurve &  1.0 & 0.96 & 0.76 & 1.0 & 1.0 \\
   \hline
\end{tabular}
}
\label{tab:hit}
\end{table}

We chose six time series datasets with diverse characteristics from the archive. The properties of the data are shown in Table \ref{tab:data}. The datasets originate from different application domains including medicine (ECG), 3D motion tracking (GunPoint), manufacturing (Wafer), synthetic sensor data (Trace) and astronomy (StarLightCurve), with different instance lengths (ranging from 82 to more than 1000).

\begin{figure}[ht]
 \centering
 \includegraphics[width=75mm]{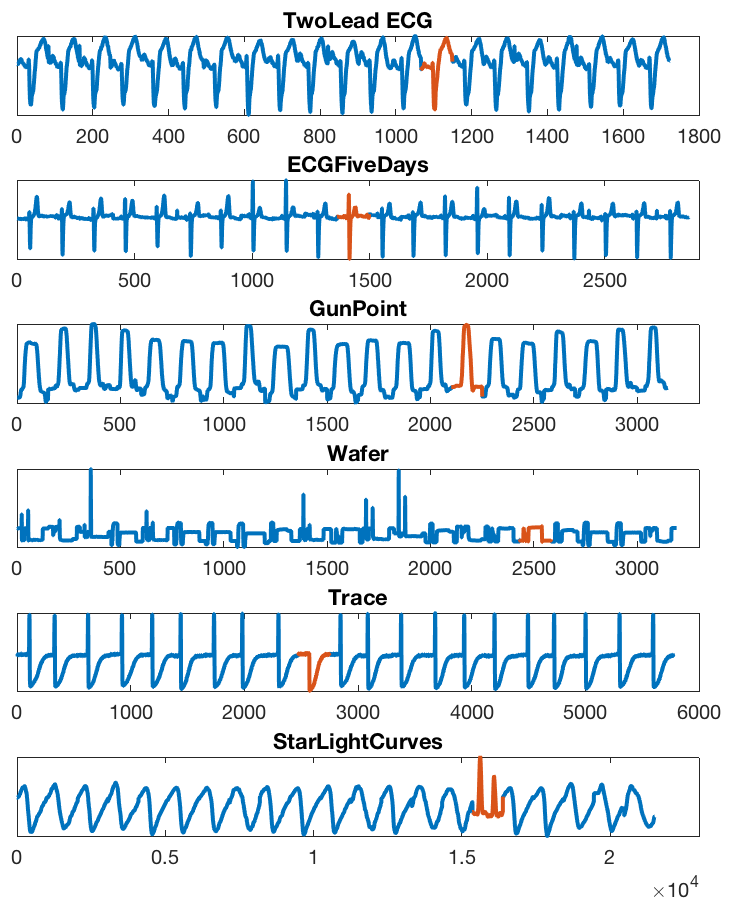}
 \caption{Examples of the test time series generated from six real-world and synthetic time series datasets. In each time series, the segment that belongs to a different class is highlighted in red.}
 \label{fig:example}
\end{figure}

The instances are labeled with class information. For each dataset, we treat all instances that belong to the first class as ``normal'' data, and all the remaining instances that belong to other classes as ``anomalous.'' We first generate a normal time series by concatenating 20 randomly selected normal instances. Then an anomalous instance is randomly selected and planted into the generated normal time series at a random position between 40\% and 80\% of the series. In this manner, we generate 25 time series for each of the six datasets. One example of the generated time series for each of the six datasets are shown in Figure \ref{fig:example}, with planted anomalous instances highlighted in red. All methods being compared are run to locate the planted anomalous instance in each time series, with sliding window length $n$ equal to the instance length.   

\subsubsection{Performance Evaluation}

Each of the tested methods returns top-3 ranked anomaly candidates (which are required to not overlap with each other) for each time series. We evaluate the performance of each method by quantifying the overlap between the discovered anomalies and the ground truth (the planted anomaly). Specifically, for each anomaly candidate, we compute a quantity called Score, defined as: 
\begin{equation}
\label{eq:score}
    \text{Score} = 1 - \min \left( 1,\frac{|\text{PredictLocation}-\text{GTLocation}|}{\text{GTLength}} \right) ,
\end{equation}
where PredictLocation is the location of the anomaly candidate, while GTLocation and GTLength are the location and length of the planted anomalous instance. The maximum value, $\text{Score} = 1$, is obtained when PredictLocation exactly matches the ground truth. The better the overlap between the anomaly candidate and the ground truth, the higher the Score. If the anomaly candidate does not overlap with the ground truth at all, then $\text{Score} = 0$. 

For each method and each time series, we only use the maximum Score achieved among the three anomaly candidates. For each method, we  record the average Score value computed over the set of 25 time series generated per dataset; the average Score values are reported in Table~\ref{tab:score}. We also record the quantity called HitRate, which is the fraction of the anomaly candidates that overlap with the ground truth (i.e., satisfy the condition $\text{Score} > 0$); the HitRate values are reported in Table~\ref{tab:hit}. Finally, we record the number of times the  ensemble grammar induction method wins/ties/loses against a baseline method; these results are reported in Table~\ref{tab:comp}.

\subsubsection{Baselines}

The proposed ensemble-based method is compared against four different baseline methods, which are described below.

\begin{itemize}
    \item \textbf{Grammar Induction with Random Parameter Values} (GI-Random): The grammar-induction-based anomaly detection approach with randomly selected parameter values $w$ and $a$. The ranges from which $w$ and $a$ are chosen are the same as in the ensemble-based approach..
    
    \item \textbf{Grammar Induction with Fixed Generic Parameter Values} (GI-Fix): The grammar-induction-based anomaly detection approach with fixed parameter values $w=4$ and $a=4$. These are popular parameter values that can be used in most of datasets as reported in~\cite{senin2014grammarviz}.
    
    \item \textbf{Grammar Induction with Selected Parameter Values} (GI-Select): The grammar-induction-based anomaly detection approach with $w$ and $a$ values selected via an optimization procedure described in~\cite{senin2018grammarviz}, using 10\% of the normal time series. The ranges from which $w$ and $a$ are chosen are the same as in the ensemble-based approach.
    
    \item \textbf{Time Series Discord} (Discord): The state-of-the-art approach that computes the one--nearest-neighbor (1-NN) distance for every subsequence in the time series; the subsequences with the largest 1-NN distances are labeled as anomalies. In the experiments, we use the latest Matrix-Profile-based implementation of this approach \cite{zhumatrix} to compute the 1-NN distances.   
\end{itemize}

\begin{table*}[t]
\caption{Wins/ties/losses of ensemble grammar induction against all baselines}
\centering
\vspace{-3mm}
\scalebox{0.8}{
\begin{tabular}{ |c||c|c|c|c|c|c| }
  \hline
  Approach \ Dataset   & TwoLeadECG  & ECGFiveDays & GunPoint & Wafer & Trace & StarLightCurve \\
    \hline\hline
  GI-Random &12/5/8 & 17/3/5 & 14/5/6 & 13/5/7 & 20/1/4 & 18/1/6\\
  GI-Fix & 17/7/1 & 13/5/7 & 15/4/6 & 17/6/2 & 14/1/10 & 24/0/1\\
  GI-Select  & 14/5/6 & 18/5/2 & 16/8/1 & 9/8/8 & 14/3/8 &17/0/8\\
  Discord  & 8/4/13 & 9/1/15 & 14/7/4 & 12/5/8 & 18/1/6 & 12/0/13\\
   \hline
\end{tabular}
}
\label{tab:comp}
\end{table*}

\begin{table*}[t]
\caption{Wins/ties/losses of ensemble grammar induction against best GI baseline for different values of $a_{\mathrm{max}}$ and $w_{\mathrm{max}}$}
\vspace{-3mm}
\centering
\scalebox{0.8}{
\begin{tabular}{ |c||c|c|c|c|c|c| }
  \hline
  Approach   & TwoLeadECG  & ECGFiveDays & GunPoint & Wafer & Trace & StarLightCurve \\
    \hline\hline
  $a_{\mathrm{max}}=5$, $w_{\mathrm{max}}=5$ &  1/12/12 & 8/9/8 & 3/9/13 &  3/14/9& 4/11/10  & 2/0/23 \\
  $a_{\mathrm{max}}=10$, $w_{\mathrm{max}}=10$ & 12/5/8 & 13/5/7 & 14/5/6 & 9/8/8 & 14/1/10 &17/0/8\\
  $a_{\mathrm{max}}=15$, $w_{\mathrm{max}}=15$ & 14/4/7  & 17/2/6 & 13/4/8& 13/7/5 &  15/0/10 & 18/0/7\\
  $a_{\mathrm{max}}=20$, $w_{\mathrm{max}}=20$ & 12/4/9&  17/2/6 & 13/4/8 & 13/7/5 & 15/0/10& 17/1/7 \\
   \hline
\end{tabular}
}
\label{tab:pa}
\end{table*}

\begin{table*}[t]
\caption{Wins/ties/losses of ensemble grammar induction against best GI baseline for different values of $w_{\mathrm{max}}$}
\centering
\vspace{-3mm}
\scalebox{0.8}{
\begin{tabular}{ |c||c|c|c|c|c|c| }
  \hline
  Approach  & TwoLeadECG  & ECGFiveDays & GunPoint & Wafer & Trace & StarLightCurve \\
    \hline\hline
  $a_{\mathrm{max}}=10$, $w_{\mathrm{max}}=5$ & 5/9/11 & 6/8/11 & 5/6/14 & 7/9/9& 4/10/11 &1/0/24 \\
  $a_{\mathrm{max}}=10$, $w_{\mathrm{max}}=10$ & 12/5/8 & 13/5/7 & 14/5/6 & 9/8/8 & 14/1/10 &17/0/8\\
  $a_{\mathrm{max}}=10$, $w_{\mathrm{max}}=15$ & 10/5/10  & 18/3/4 & 11/6/8 & 18/3/4 & 15/0/10 & 19/0/6\\
  $a_{\mathrm{max}}=10$, $w_{\mathrm{max}}=20$  & 12/4/9  & 18/2/5 & 10/4/11 & 14/3/8 & 16/0/9 & 20/0/5\\
   \hline
\end{tabular}
}
\label{tab:p}
\end{table*}

\begin{table*}[t]
\caption{Wins/ties/losses of ensemble grammar induction against best GI baseline for different values of $a_{\mathrm{max}}$}
\centering
\vspace{-3mm}
\scalebox{0.8}{
\begin{tabular}{ |c||c|c|c|c|c|c| }
  \hline
  Approach   & TwoLeadECG  & ECGFiveDays & GunPoint & Wafer & Trace & StarLightCurve \\
    \hline\hline
  $a_{\mathrm{max}}=5$, $w_{\mathrm{max}}=10$ & 11/5/9  & 8/8/9 & 7/8/10 & 12/7/6 & 11/5/9 & 1/1/23\\
  $a_{\mathrm{max}}=10$, $w_{\mathrm{max}}=10$ & 12/5/8 & 13/5/7 & 14/5/6 & 9/8/8 & 14/1/10 &17/0/8\\
  $a_{\mathrm{max}}=15$, $w_{\mathrm{max}}=10$ & 11/6/8  & 13/6/6 & 13/4/8 & 8/8/9 & 16/0/9 & 15/0/10\\
  $a_{\mathrm{max}}=20$, $w_{\mathrm{max}}=10$  & 11/4/10  & 14/5/6 & 13/4/8 & 9/9/7 & 15/0/10 & 12/1/12\\
   \hline
\end{tabular}
}
\label{tab:a}
\end{table*}

\subsubsection{Results}

The overall performance results measured by the average Score and HitRate are shown in Tables~\ref{tab:score}~and~\ref{tab:hit}, respectively. 

According to Table~\ref{tab:score}, the proposed approach achieves the highest average Score values in four out of six datasets, and the second-highest average Score values in two of the datasets. Compared with GI-Random, GI-Fix, and GI-Select, the proposed approach achieves higher average Score values in all six datasets. Compared with Discord, the proposed approach achieves a higher average Score value in four out of six datasets (GunPoint, Wafer, Trace and StarLightCurve). In two datasets (ECGFiveDays and TwoLeadECG), Discord outperforms the proposed approach. 

According to Table~\ref{tab:hit}, the proposed approach achieves the highest HitRate values (alone or shared) in five out of six datasets, and the second-highest HitRate value in one of the datasets. These results are consistent with those in Table \ref{tab:score} and indicate that the proposed approach can successfully locate an anomalous subsequence that strongly overlaps with the ground truth.

\begin{figure*}[t]
    \centering
    \begin{subfigure}[b]{0.25\textwidth}
        \includegraphics[width=\textwidth]{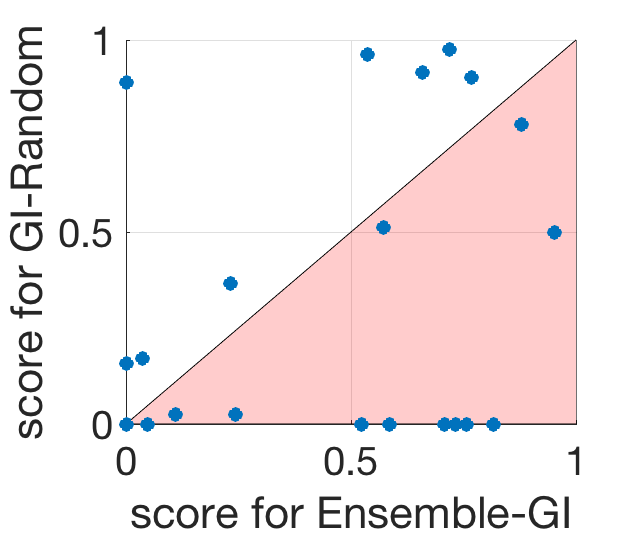}
        \caption{Vs. GI-Random (TwoLeadECG)}
        \label{fig:result-a}
    \end{subfigure}
    ~ 
    \begin{subfigure}[b]{0.25\textwidth}
        \includegraphics[width=\textwidth]{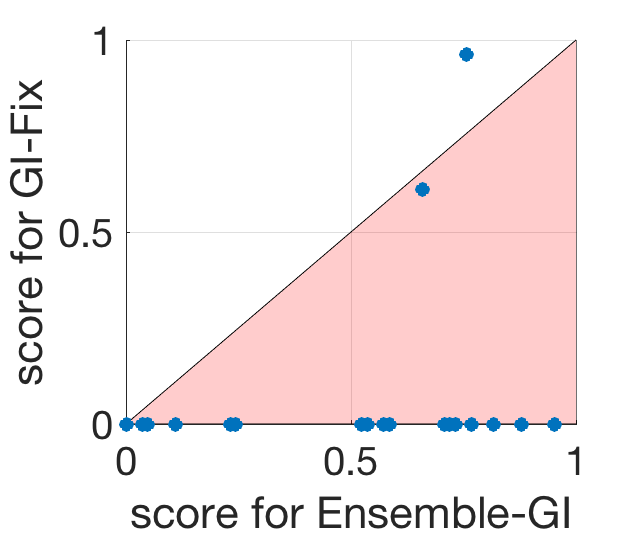}
        \caption{Vs. GI-Fix (TwoLeadECG)}
        \label{fig:result-b}
    \end{subfigure}
    ~ 
    \begin{subfigure}[b]{0.25\textwidth}
        \includegraphics[width=\textwidth]{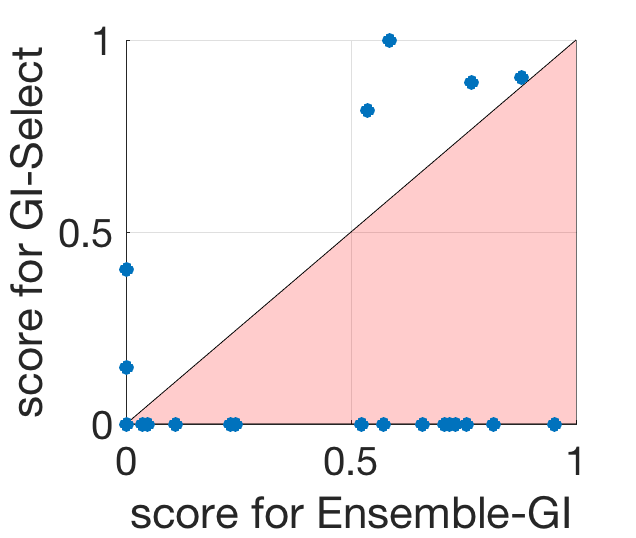}
        \caption{Vs. GI-Select (TwoLeadECG)}
        \label{fig:result-c}
    \end{subfigure}
    ~
       \begin{subfigure}[b]{0.25\textwidth}
        \includegraphics[width=\textwidth]{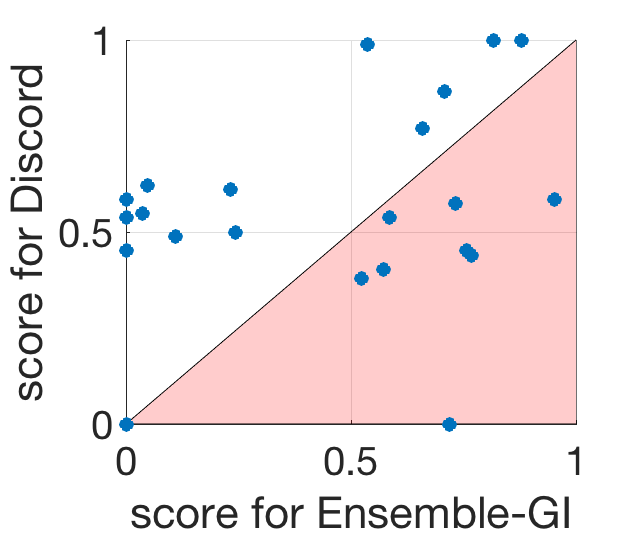}
        \caption{Vs. Discord (TwoLeadECG)}
        \label{fig:result-d}
    \end{subfigure}
    
       \begin{subfigure}[b]{0.25\textwidth}
        \includegraphics[width=\textwidth]{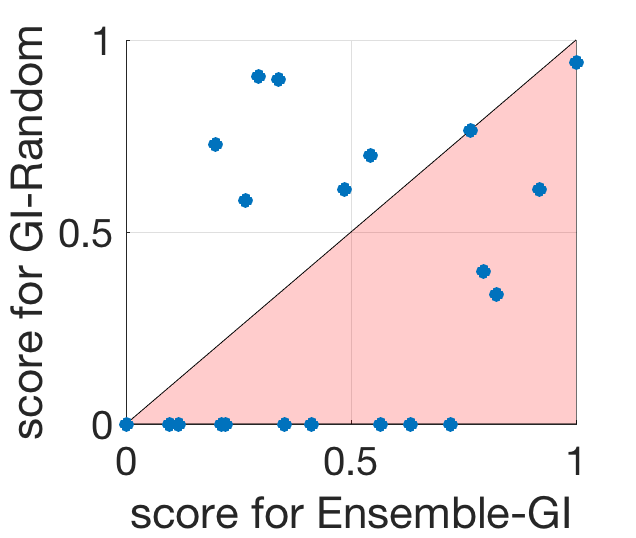}
        \caption{Vs. GI-Random (ECGFiveDays)}
        \label{fig:result-e}
    \end{subfigure}
    ~
       \begin{subfigure}[b]{0.25\textwidth}
        \includegraphics[width=\textwidth]{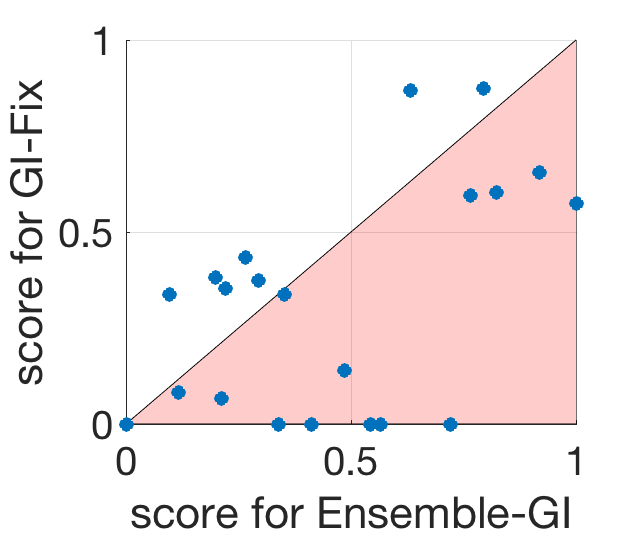}
        \caption{Vs. GI-Fix (ECGFiveDays)}
        \label{fig:result-f}
       \end{subfigure} 
    ~   
      \begin{subfigure}[b]{0.25\textwidth}
        \includegraphics[width=\textwidth]{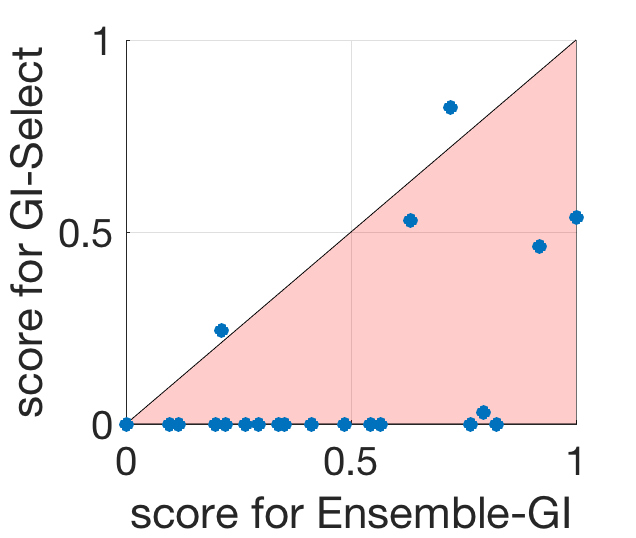}
        \caption{Vs. GI-Select (ECGFiveDays)}
        \label{fig:result-g}
    \end{subfigure}
    ~
       \begin{subfigure}[b]{0.25\textwidth}
        \includegraphics[width=\textwidth]{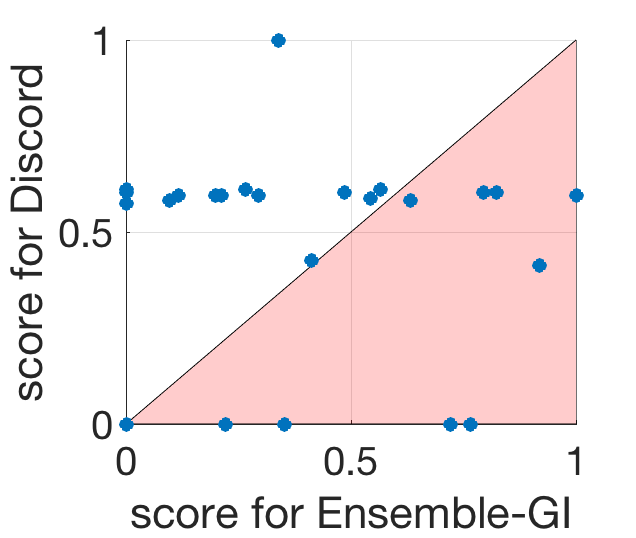}
        \caption{Vs. Discord (ECGFiveDays)}
        \label{fig:result-h}
    \end{subfigure}
    
       \begin{subfigure}[b]{0.25\textwidth}
        \includegraphics[width=\textwidth]{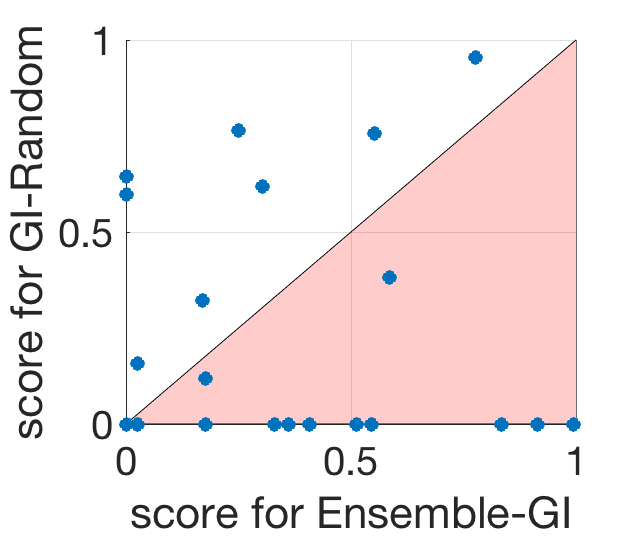}
        \caption{Vs. GI-Random (Wafer)}
        \label{fig:result-i}  
    \end{subfigure}
    ~
    \begin{subfigure}[b]{0.25\textwidth}
        \includegraphics[width=\textwidth]{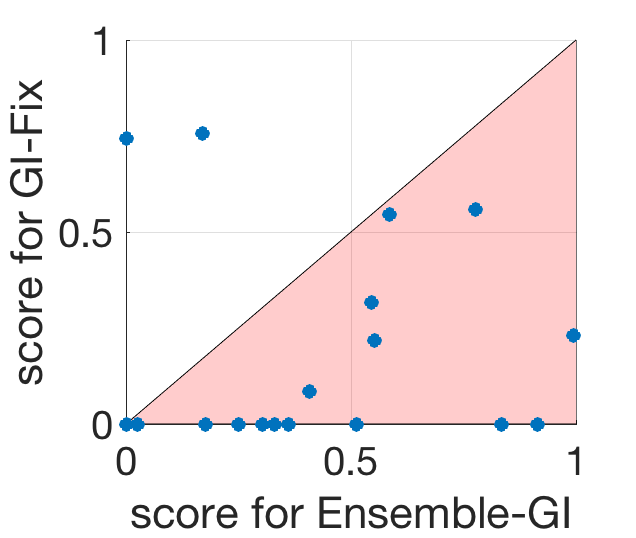}
        \caption{Vs. GI-Fix (Wafer)}
        \label{fig:result-j}  
    \end{subfigure}
    ~
    \begin{subfigure}[b]{0.25\textwidth}
        \includegraphics[width=\textwidth]{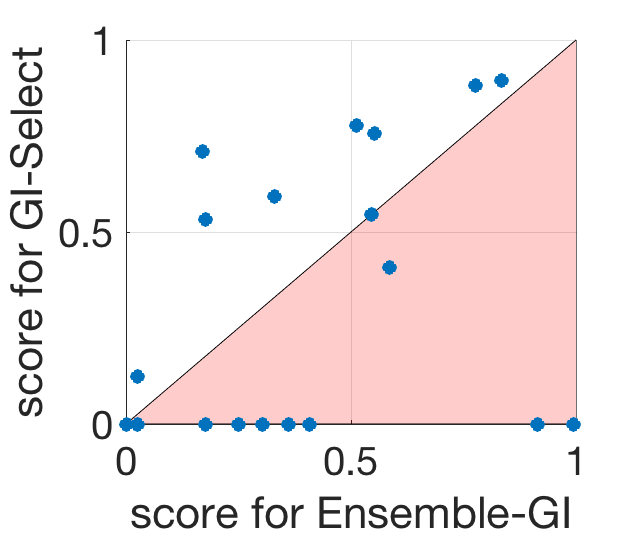}
        \caption{Vs. GI-Select (Wafer)}
        \label{fig:result-k}  
    \end{subfigure}
    ~
    \begin{subfigure}[b]{0.25\textwidth}
        \includegraphics[width=\textwidth]{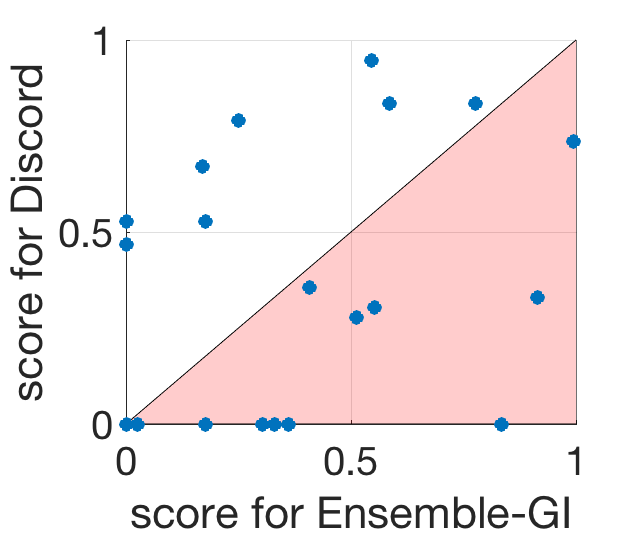}
        \caption{Vs. Discord (Wafer)}
        \label{fig:result-l}  
    \end{subfigure}
    
    \begin{subfigure}[b]{0.25\textwidth}
        \includegraphics[width=\textwidth]{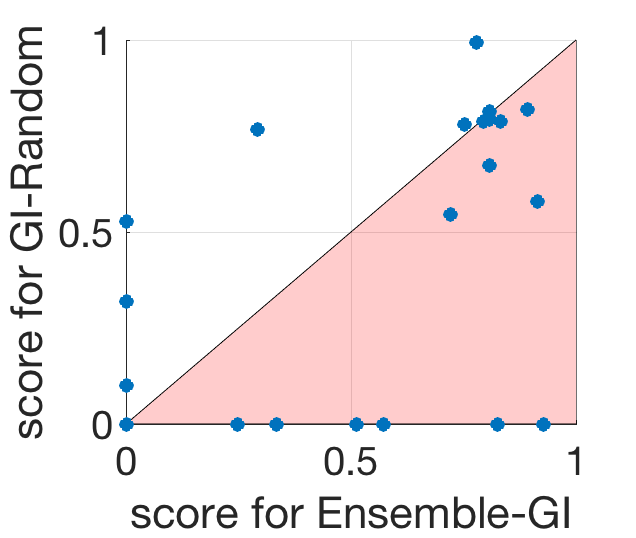}
        \caption{Vs. GI-Random (GunPoint)}
        \label{fig:result-m}  
    \end{subfigure}
    ~
    \begin{subfigure}[b]{0.25\textwidth}
        \includegraphics[width=\textwidth]{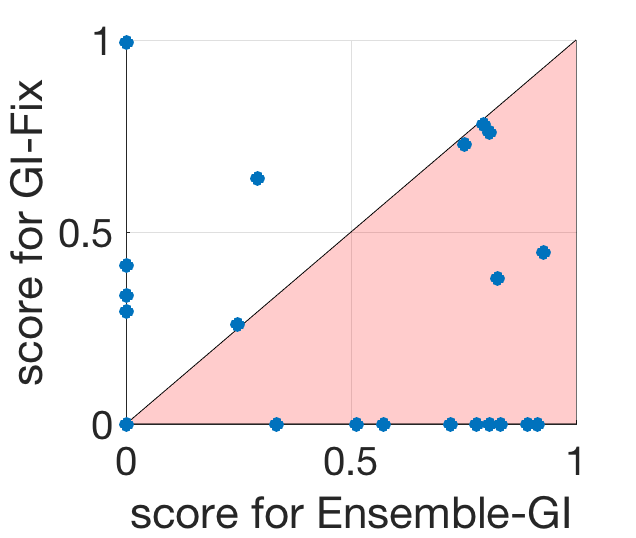}
        \caption{Vs. GI-Fix (GunPoint)}
        \label{fig:result-n}  
    \end{subfigure}
    ~
    \begin{subfigure}[b]{0.25\textwidth}
        \includegraphics[width=\textwidth]{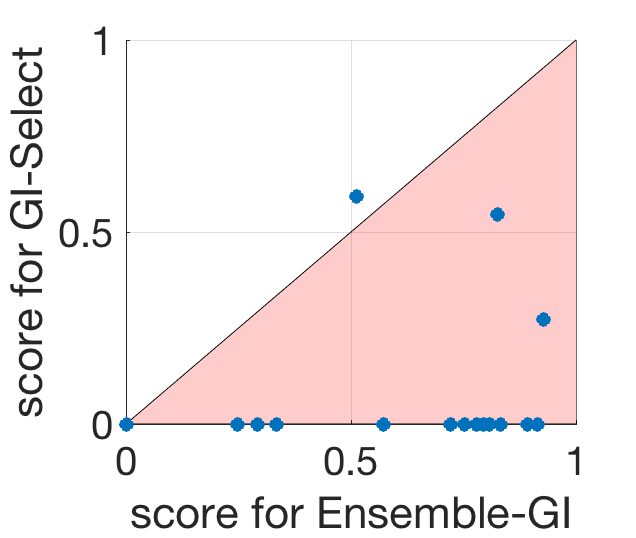}
        \caption{Vs. GI-Select (GunPoint)}
        \label{fig:result-o}  
    \end{subfigure}
    ~
    \begin{subfigure}[b]{0.25\textwidth}
        \includegraphics[width=\textwidth]{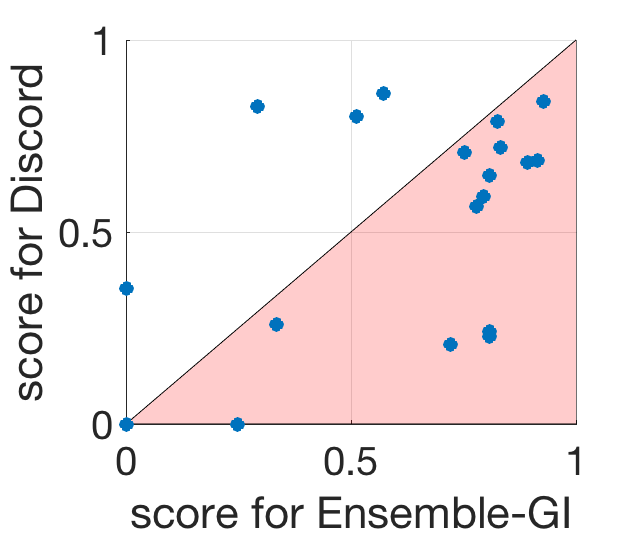}
        \caption{Vs. Discord (GunPoint)}
        \label{fig:result-p}  
    \end{subfigure}

       \begin{subfigure}[b]{0.25\textwidth}
        \includegraphics[width=\textwidth]{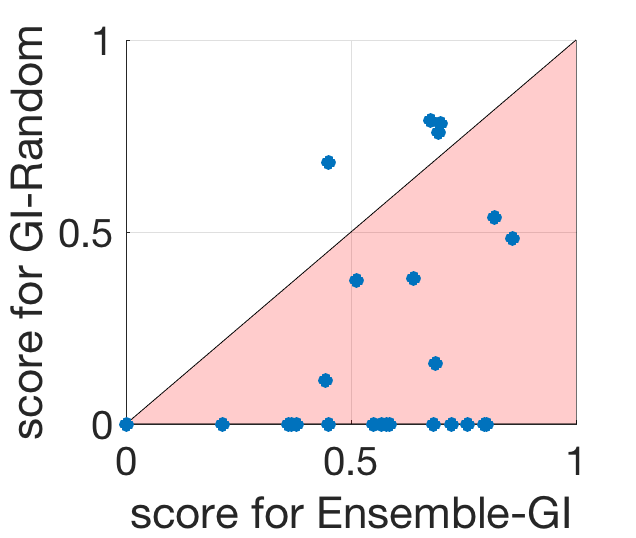}
        \caption{Vs. GI-Random (Trace)}
        \label{fig:result-q}  
    \end{subfigure}
    ~
    \begin{subfigure}[b]{0.25\textwidth}
        \includegraphics[width=\textwidth]{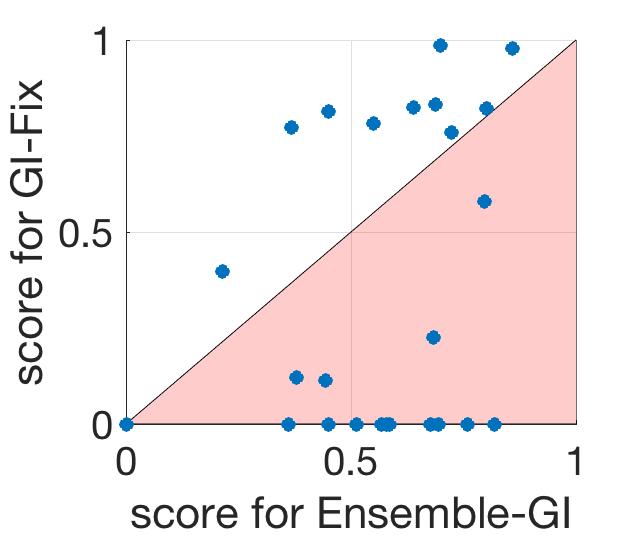}
        \caption{Vs. GI-Fix (Trace)}
        \label{fig:result-r}  
    \end{subfigure}
    ~
    \begin{subfigure}[b]{0.25\textwidth}
        \includegraphics[width=\textwidth]{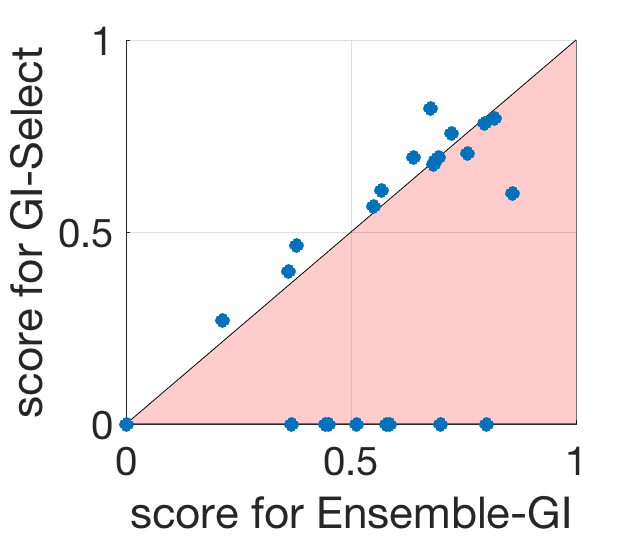}
        \caption{Vs. GI-Select (Trace)}
        \label{fig:result-s}  
    \end{subfigure}
    ~
    \begin{subfigure}[b]{0.25\textwidth}
        \includegraphics[width=\textwidth]{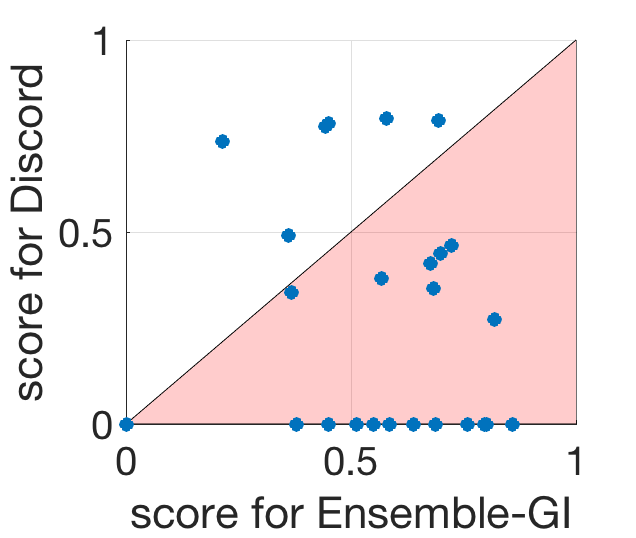}
        \caption{Vs. Discord (Trace)}
        \label{fig:result-t}  
    \end{subfigure}
    
         \begin{subfigure}[b]{0.26\textwidth}
        \includegraphics[width=\textwidth]{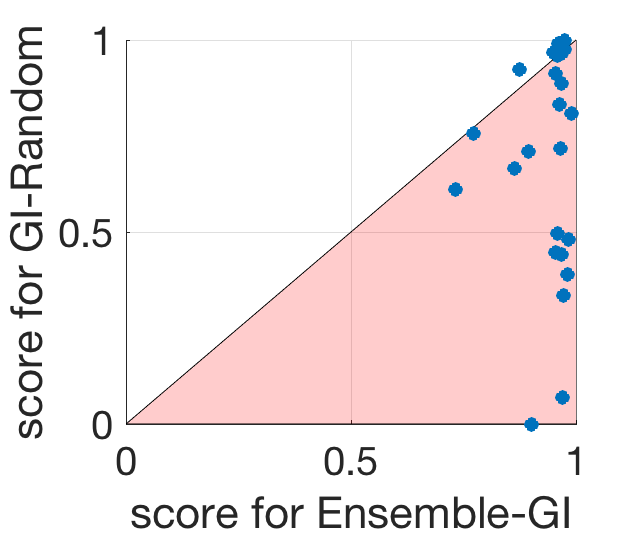}
        \caption{Vs. GI-Random (StarLightCurve)}
        \label{fig:result-u}  
    \end{subfigure}
    ~
    \begin{subfigure}[b]{0.25\textwidth}
        \includegraphics[width=\textwidth]{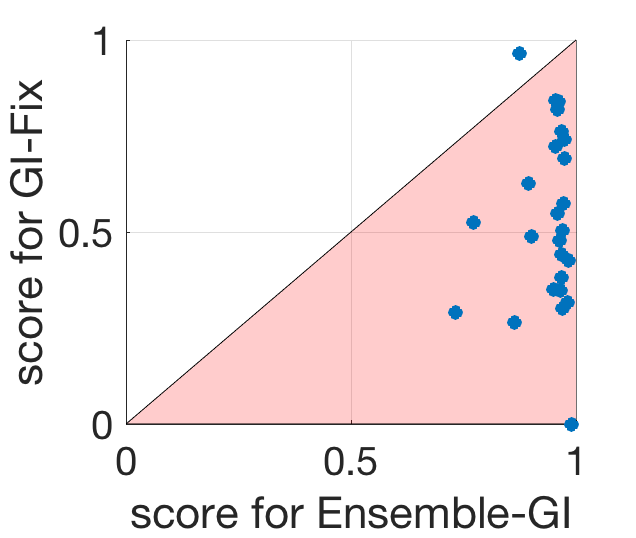}
        \caption{Vs. GI-Fix (StarLightCurve)}
        \label{fig:result-v}  
    \end{subfigure}
    ~
    \begin{subfigure}[b]{0.25\textwidth}
        \includegraphics[width=\textwidth]{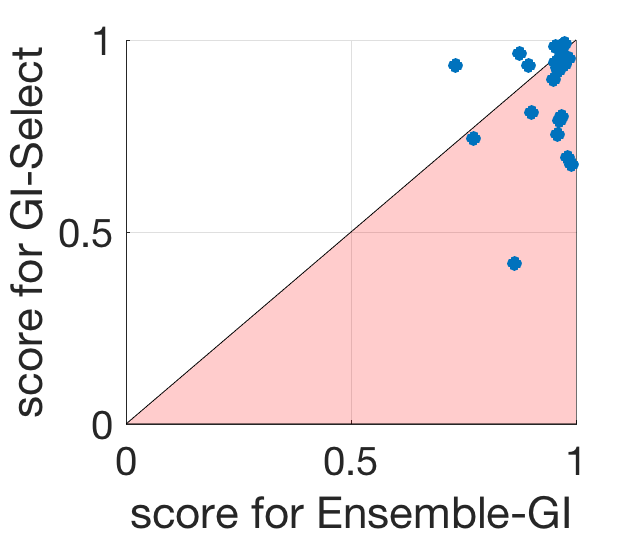}
        \caption{Vs. GI-Select (StarLightCurve)}
        \label{fig:result-w}  
    \end{subfigure}
    ~
    \begin{subfigure}[b]{0.25\textwidth}
        \includegraphics[width=\textwidth]{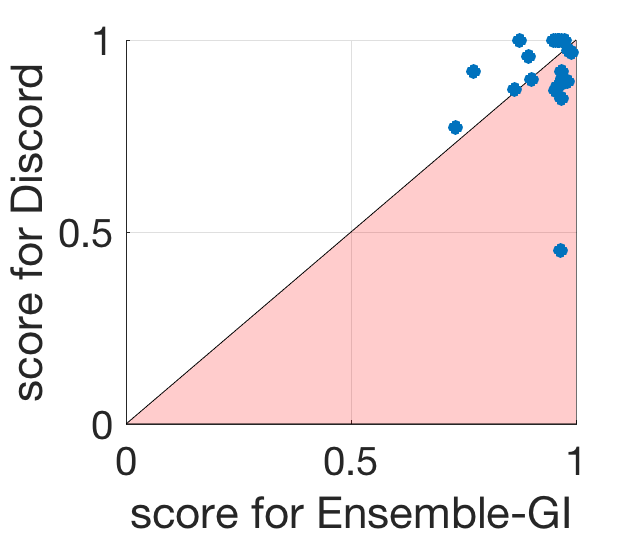}
        \caption{Vs. Discord (StarLightCurve)}
        \label{fig:result-x}  
    \end{subfigure}
    \caption{Summary of performance comparison of ensemble grammar induction against baseline methods.}
    \label{fig:result}

\end{figure*}

In addition, Figure~\ref{fig:result} shows a detailed comparison summary of the proposed approach against all baselines. Each of the six rows corresponds to one of the datasets, and each of the four columns corresponds to one of the baselines. In each plot, a blue dot located at the point $(x, y)$ denotes a pair of Score values (ensemble Score, baseline Score), computed from Eq.~\eqref{eq:score} for one of the 25 generated time series. A dot located in the lower triangle (highlighted in pink) of the plot corresponds to a win of the proposed approach (ensemble Score $>$ baseline Score); a point located in the upper triangle corresponds to a loss (ensemble Score $<$ baseline Score), and a point located on the diagonal corresponds to a tie (ensemble Score $=$ baseline Score). The numbers of wins, ties, and losses of the proposed method against the baselines are shown in Table~\ref{tab:comp}. 

According to Figure \ref{fig:result}, the proposed approach outperforms GI-Random, GI-Fix, and GI-Select. Moreover, in most of the tested datasets, the proposed method often detects ground truth anomalies that are completely missed by these variations of the grammar-induction-based approach (i.e., the respective baseline Score values are zero), while the opposite outcomes (ensemble Score values are zero) are quite rare. Compared with Discord, the proposed approach achieves similar performance. Moreover, in most of the tested datasets, cases where the ensemble-based approach discovers ground truth anomalies missed by Discord are more common than the opposite ones.

According to Table~\ref{tab:comp}, compared with GI-Random, GI-Fix, and GI-Select, the proposed method wins in more than half of the time series in most datasets. Compared with Discord, the results are similar to those measured by Score and HitRate. Specifically, the proposed approach has more wins than losses in three datasets (GunPoint, Wafer, and Trace), loses more often than wins in two ECG datasets, and is in a virtual dead heat with Discord in the StarLightCurves dataset.

In summary, the experiments indicate that using an ensemble of parameter values can significantly improve the performance of the grammar-induction-based approach, compared to the variations of the method where a single combination of parameter values is selected. Also, the proposed method can achieve a competitive performance compared with the state-of-the-art discord approach. Importantly, while the latter has a quadratic time complexity, the former preserves a linear time complexity and hence is much more feasible for anomaly detection in large-scale data.

\subsection{Effects of Parameter Value Ranges}

In this subsection, we first evaluate how the ensemble grammar induction approach performs for different parameter value ranges determined by $w_{\mathrm{max}}$ and $a_{\mathrm{max}}$. Specifically, we evaluate the performance on the same time series that were used in Sec. 7.1. As the baseline for comparison, we use the best of the GI-Random, GI-Fix, and GI-Select methods for each dataset. We report the number of wins/ties/loses vs. the baseline for all tested parameter value ranges. We then evaluate how values of three hyper parameters --- ensemble size $N$, ensemble selectivity $\tau$, and sliding window length $n$ --- affect the result. In these experiments, we evaluate the performance based on Score and HitRate.

\subsubsection{Effect of $w_{\mathrm{max}}$ and $a_{\mathrm{max}}$}
We first test the proposed approach with values of $w_{\mathrm{max}}$ and $a_{\mathrm{max}}$ equal to each other and varying from 5 to 20.  The results are shown in Table~\ref{tab:pa}. According to the results, the smallest ranges ($w_{\mathrm{max}}= a_{\mathrm{max}} = 5$) lead to the worst performance. A possible explanation is that the ranges are too small to generate a sufficient number of high-quality rule density curves. The performance significantly improves for $w_{\mathrm{max}} = a_{\mathrm{max}} = 15$. However, increasing the range values beyond 15 is not useful; furthermore, in two datasets (TwoLeadECG and StarLightCurve) the performance slightly deteriorates when the ranges go up to 20, and in the GunPoint dataset the performance actually peaks for $w_{\mathrm{max}} = a_{\mathrm{max}} = 10$. The lack of performance improvement (or even some deterioration) at range values larger than 15 indicates that long and high-resolution SAX words may capture too much noise in data and, as a result, produce low-quality rule density curves. Nevertheless, the ensemble-based method still outperforms other approaches that rely on a single combination of parameter values.

\subsubsection{Effect of $w_{\mathrm{max}}$}

We also test the proposed approach with values of $w_{\mathrm{max}}$ varying from 5 to 20 and fixed $a_{\mathrm{max}}=10$. The results are shown in Table~\ref{tab:p}. Once again, we observe that when the range is too small ($w_{\mathrm{max}} = 5$), the performance is the worst. The performance significantly improves for larger values of $w_{\mathrm{max}}$, with the peak-performance value depending on the dataset.

\subsubsection{Effect of $a_{\mathrm{max}}$}

In this experiment, we test the proposed approach with values of $a_{\mathrm{max}}$ varying from 5 to 20 and fixed $w_{\mathrm{max}}=10$. The results are shown in Table~\ref{tab:a}. For $a_{\mathrm{max}} = 5$, the performance is subpar for the ECGFiveDays, GunPoint, and Trace datasets, and it is terrible for the StarLightCurve dataset (as a rule, this dataset exhibits the worst performance when one or both range values are small). However, $a_{\mathrm{max}} = 5$ actually results in the best performance for the Wafer dataset. For most datasets, the $a_{\mathrm{max}}$ values of 10, 15, and 20 produce very similar results.

The last two experiments indicate that having a larger range for $w$ is more important than for $a$, which may indicate that a variation in the PAA size has a larger effect on the performance of the algorithm than a variation in the alphabet size, and hence $w$ is a more important parameter to choose, in general.


\begin{table}[ht]
\caption{Performance (average Score) vs. $N$}
\vspace{-3mm}
\centering
\scalebox{0.8}{
\begin{tabular}{|c|c|c|c|c|} 
\hline
      Dataset & $N=5$ & $N=10$ & $N=25$ & $N=50$ \\
     \hline\hline
	TwoLeadECG & 0.3424 & 0.3488 & 0.3912 & 0.3951\\ 
	ECGFiveDays & 0.37 & 0.3882 & 0.4168 & 0.3903 \\ 
	GunPoint &  0.3128 & 0.4629  &  0.4965 & 0.4728\\ 
	Wafer & 0.2308 &   0.2637  &  0.2839 & 0.3179\\ 
	Trace &  0.4767  &  0.5789  &  0.5994 & 0.5718\\ 
	StarLightCurve & 0.8244 & 0.7593 & 0.8676 & 0.9369\\ 
	\hline
\end{tabular}
}
\label{tab:nscore}
\end{table}

\begin{table}[ht]
\caption{Performance (HitRate) vs. $N$}
\centering
\scalebox{0.8}{
\begin{tabular}{|c|c|c|c|c|} 
\hline
     Dataset & $N=5$ & $N=10$ & $N=25$ & $N=50$ \\
     \hline\hline
	TwoLeadECG & 0.52 & 0.6 & 0.72 & 0.72\\ 
	ECGFiveDays  & 0.68 & 0.72 & 0.76 & 0.8\\ 
	GunPoint & 0.56 & 0.76 & 0.68 & 0.68\\ 
	Wafer &0.44 & 0.64 & 0.6 & 0.72\\ 
	Trace &0.76 & 0.96 & 0.96 & 0.96\\ 
	StarLightCurve &1.0 & 1.0 & 1.0 & 1.0\\
	\hline
\end{tabular}
}
\label{tab:nhit}
\end{table}

\subsubsection{Effect of Ensemble Size}

We next evaluate the proposed method with the ensemble size $N$ varying from 5 to 50 (recall that all other experiments used the fixed value $N = 50$). The average Score and HitRate of the proposed method for different ensemble sizes are shown in Tables \ref{tab:nscore} and \ref{tab:nhit}, respectively.

We observe that the performance for a small ensemble size ($N=5$)  is worse than for a larger ensemble size ($N=25$ or $N=50$) for all datasets. In general, the Score and HitRate values increase as $N$ grows, but these increases saturate when the ensemble size is large enough (e.g., when $N \geq 25$). The results indicate that selecting $N \geq 25$ may be suitable for most cases.

We also observe that the Score values are, in general, rather low for all methods, for most datasets except StarLightCurve. This could be due to the fact that the Score is normalized by the ground truth length (cf.~Eq.~\eqref{eq:score}). However, the planted anomaly may only have a small section that differs from the ``normal'' data, and that may be what the algorithms detect. In such case, treating the entire planted instance as anomalous would indeed lower the Score. This explanation is later validated by the results in Table~\ref{tab:len_p}, which show that a shorter sliding window length results in higher Score values in some cases. Nevertheless, this is why we also use the HitRate, which is independent of the ground truth length, as an alternate measure.

\subsubsection{Effects of Ensemble Selectivity}

Next, we evaluate the proposed method with ensemble selectivity $\tau$ varying from 5\% to 100\%. In this experiment, the evaluation of the average Score value (which is computed over 25 time series) is repeated 20 times for each dataset and $\tau$ value. We then compute the mean and standard deviation over the set of 20 average Score values. The results are shown in Table~\ref{tab:tau}. We observe that better performance is typically achieved for smaller $\tau$ values (e.g., from 5\% for ECGFiveDays and Trace datasets to 20\% for GunPoint and StarLightCurve datasets), while the standard deviation is typically smaller for $\tau$ values of 20\% or 40\%. Therefore, we recommend choosing $\tau$ around 20\% since it combines a relatively high level of performance with a relatively small variance of results.

\begin{table}[ht]
\caption{Mean and standard deviation over 20 average Score values, vs. $\tau$}
\vspace{-2mm}
\centering
\scalebox{0.75}{
\begin{tabular}{|c|c|c|c|c|c|c|} 
\hline
     Dataset & $\tau=5\%$ & $\tau=10\%$ & $\tau=20\%$ & $\tau=40\%$ & $\tau=80\%$ & $\tau=100\%$\\
     
     \hline\hline
	TwoLeadECG & 0.4149 & 0.4196 & 0.4& 0.3882 & 0.3354 & 0.3071\\ 
	 & (0.04) & (0.032) & (0.026)& (0.027) & (0.036) & (0.032)\\
	 \hline
	ECGFiveDays  & 0.425 & 0.41 & 0.38 & 0.37 & 0.35& 0.32\\ 
	& (0.042) & (0.045) & (0.038)& (0.037) & (0.024) & (0.036)\\
		 \hline
	GunPoint & 0.488 & 0.50 & 0.505 & 0.488 & 0.43 & 0.412\\ 
	& (0.042) & (0.037) & (0.035)& (0.025) & (0.023) & (0.023)\\
		 \hline
	Wafer &0.339 & 0.371 & 0.337 & 0.311 & 0.27 & 0.26\\ 
	& (0.05) & (0.042) & (0.027)& (0.027) & (0.032) & (0.037)\\
		 \hline
	Trace &0.6136 & 0.6017 & 0.5972 & 0.5864 & 0.4997 & 0.4166\\ 
	& (0.037) & (0.035) & (0.025)& (0.024) & (0.046) & (0.042)\\
		 \hline
	StarLightCurve &0.9057 & 0.9183 & 0.9327 & 0.9052 & 0.7359 & 0.628\\
	& (0.017) & (0.016) & (0.009)& (0.012) & (0.021) & (0.021)\\
	\hline
\end{tabular}
}
\vspace{-1mm}
\label{tab:tau}
\end{table}

\begin{table}[ht]
\caption{Performance (average Score) vs. $n$}
\vspace{-2mm}
\centering
\scalebox{0.8}{
\begin{tabular}{|c|c|c|c|c|c|} 
\hline
     Dataset &$n = 0.6 n_{\mathrm{a}}$ & $n = 0.7 n_{\mathrm{a}}$ & $n = 0.8 n_{\mathrm{a}}$ & $n = 0.9 n_{\mathrm{a}}$ & $n = n_{\mathrm{a}}$ \\
     \hline\hline
	TwoLeadECG & 0.4620   & 0.4605  &  0.4107  &  0.4259 & 0.3951\\ 
	ECGFiveDays  & 0.4391   & 0.3691   & 0.3535   & 0.3797 & 0.3903\\ 
	GunPoint &0.4373   & 0.4992 &   0.4680 &   0.4371 & 0.4728\\ 
	Wafer &0.3095  &  0.4195  &  0.3389  &  0.2824 & 0.3179\\ 
	Trace &0.5229   & 0.5911  &  0.5689 &   0.5852 & 0.5718\\ 
	StarLightCurve &0.8624 & 0.8998 & 0.9216 & 0.9048 & 0.9369\\
	\hline
\end{tabular}
}
\vspace{-1mm}
\label{tab:len_p}
\end{table}

\begin{table}[ht]
\caption{Performance (HitRate) vs. $n$}
\centering
\vspace{-3mm}
\scalebox{0.8}{
\begin{tabular}{|c|c|c|c|c|c|} 
\hline
     Dataset &$n = 0.6 n_{\mathrm{a}}$ & $n = 0.7 n_{\mathrm{a}}$ & $n = 0.8 n_{\mathrm{a}}$ & $n = 0.9 n_{\mathrm{a}}$ & $n = n_{\mathrm{a}}$ \\
     
     \hline\hline
	TwoLeadECG & 0.72 & 0.84 & 0.8& 0.76 & 0.72\\ 
	ECGFiveDays  & 0.96 & 0.8 & 0.84 & 0.72 & 0.8\\ 
	GunPoint & 0.84 & 0.68 & 0.72 & 0.64 & 0.68\\ 
	Wafer &0.56 & 0.64 & 0.52 & 0.52 & 0.72\\ 
	Trace &1.0 & 1.0 & 1.0 & 1.0 & 0.96\\ 
	StarLightCurve &1.0 & 1.0 & 1.0 & 1.0 & 1.0\\
	\hline
\end{tabular}
}
\vspace{-2mm}
\label{tab:len_h}
\end{table}

\subsubsection{Effects of Sliding Window Length}

In this experiment, we investigate how the proposed method performs when the sliding window length $n$ is less than the ground truth anomaly length (denoted as $n_{\mathrm{a}}$). Specifically, we test five different sliding window lengths equal to 60\%, 70\%, 80\%, 90\% and 100\% of $n_{\mathrm{a}}$. The average Score and HitRate values are shown in Tables \ref{tab:len_p} and \ref{tab:len_h}, respectively. According to the results, while there exists some variation in the performance, the dependence on $n$ is not significant, and the proposed method robustly outperforms the existing grammar-induction-based approaches in most cases.

\begin{figure}[ht]
    \centering
    \begin{subfigure}[b]{43mm}
        \includegraphics[width=\textwidth]{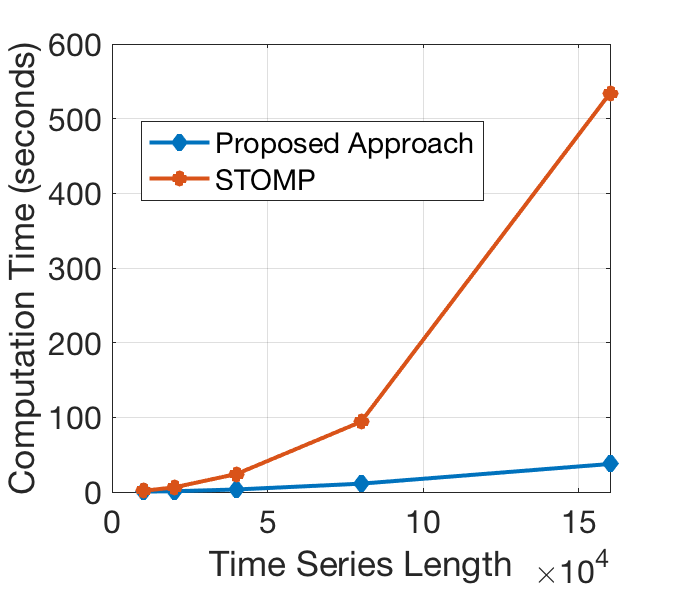}
        \caption{RW time series}
        \label{fig:scal-a}
    \end{subfigure}
    \begin{subfigure}[b]{43mm}
        \includegraphics[width=\textwidth]{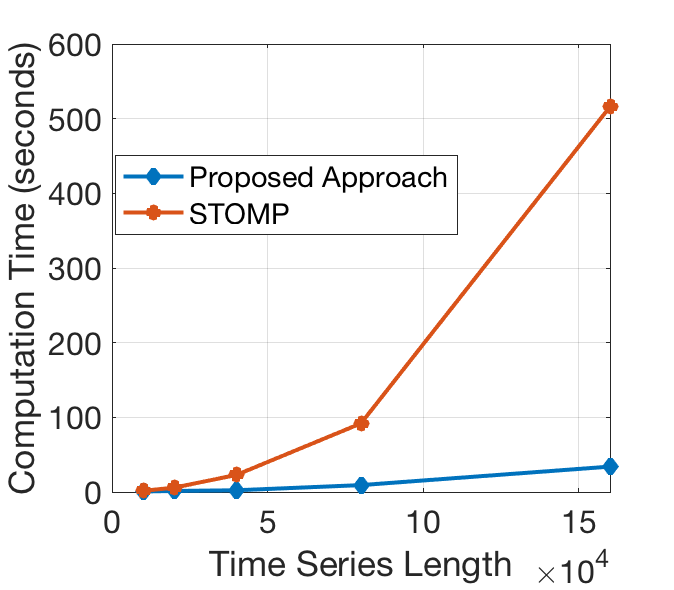}
        \caption{ECG time series}
        \label{fig:scal-b}
    \end{subfigure}
    \begin{subfigure}[b]{43mm}
        \includegraphics[width=\textwidth]{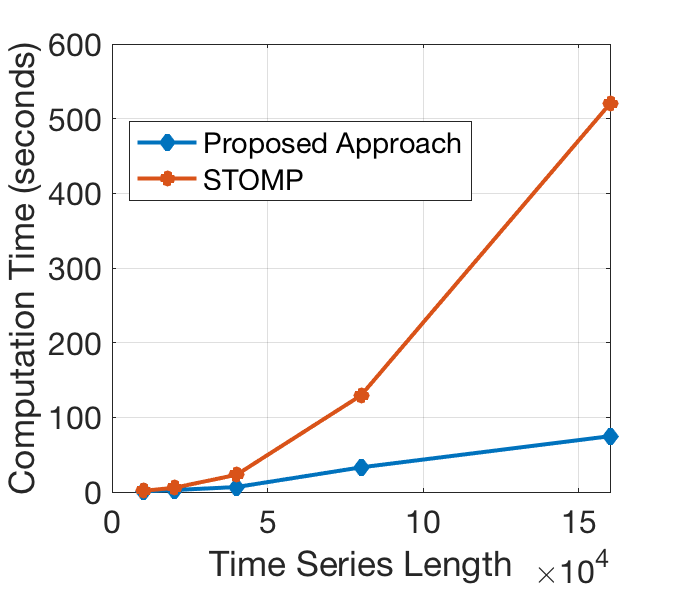}
        \caption{EEG time series}
        \label{fig:scal-c}
    \end{subfigure}
   \caption{Scalability: Computation time vs. time series length.}
   \label{fig:scal}
   \vspace{-5mm}
\end{figure}

\subsection{Scalability}

We evaluate the scalability of the proposed ensemble-based approach by applying it to a 160,000 length random walk (RW) time series, ECG data~\cite{senin2015time} and electroencephalogram (EEG) data~\cite{mueen2013enumeration}. We compare the computation times for the proposed method and for the state-of-the-art discord discovery approach. In the experiment, we use STOMP \cite{zhumatrix}, the latest Matrix-Profile-based algorithm, to detect discords. It has been shown that STOMP is both faster and more robust to different types of data compared to the original discord discovery method HOTSAX \cite{keogh2005hot}.

The scalability, evaluated as the computation time versus the time series length, is shown in Figure~\ref{fig:scal}. We see that, as the time series length increases, the computation time grows significantly slower for the proposed method than for STOMP. At the largest time series length, the proposed approach is about one order of magnitude faster than STOMP, for all three types of time series data. We also find that the computation times for both approaches are roughly independent of the sliding window length.

\subsection{Case Study: Anomaly Detection in Electric Power Usage Time Series}

\begin{figure}[ht]
 \centering
 \includegraphics[width=65mm]{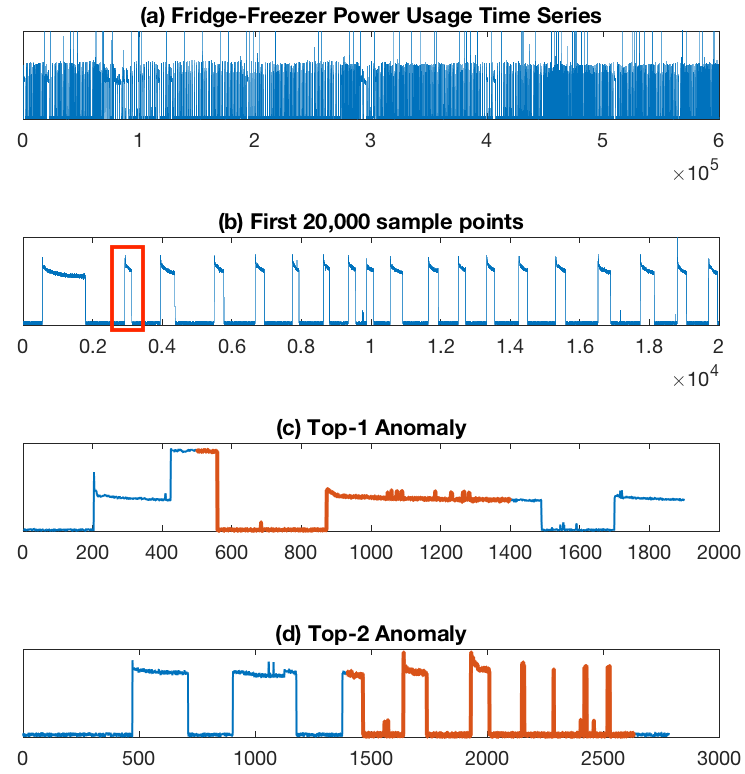}
 \caption{A 600,000 length fridge-freezer power usage time series, and two top-ranked anomalies detected in this series by ensemble grammar induction.}
 \label{fig:pu-anomaly}
\vspace{-6mm}
\end{figure}

Interpreting the time dependence of electric power usage has many potential applications \cite{yeh2016matrix}. In this section, we show that the ensemble grammar induction method can find anomalies in large-scale electric power usage time series.
We use 100 days of the fridge-freezer power usage data provided in \cite{murray2015energy}, to evaluate the performance of the proposed method. The entire time series, which consists of approximately 600,000 points, is shown in Figure \ref{fig:pu-anomaly}(a), and the first 20,000 points are shown in Figure \ref{fig:pu-anomaly}(b). We run the proposed method with the sliding window length of 900, which is about the duration of one cycle (shown in red box in Figure \ref{fig:pu-anomaly}(b)). The computation time is about one minute. Figures \ref{fig:pu-anomaly}(c) and \ref{fig:pu-anomaly}(d) show the two top-ranked anomaly candidates detected by the proposed method. 

We see that the top-1 anomaly represents a cycle whose shape is unusual compared to the typical cycles shown in Figure \ref{fig:pu-anomaly}(b). The top-2 anomaly represents an unusual event that contains normal cycles and short spikes. The two anomalies represent different, unusual power usage patterns. Since this time series is very long, and the anomalies have different lengths, using the state-of-the-art discord discovery approach would be time consuming. In contrast, the proposed method provides an efficient way to detect anomalies.

\subsection{Detecting Multiple Anomalies}

We also investigate the effectiveness of the proposed approach in detecting multiple anomalies in time series. In this experiment, we use the StarlightCurve dataset to generate 10 time series. Each of these time series is of length 43008 and contains two randomly selected and placed anomalies of length 1024. We evaluate the performance by the number of ground truth anomalies detected (a ground truth anomaly is considered detected if it overlaps with at least one of the top-3 ranked anomaly candidates). The proposed method performed well as it successfully identified both anomalies in nine time series and one of the two anomalies in one time series.

\section{Conclusion}

In this paper, we introduce a robust grammar-induction-based anomaly detection approach utilizing ensemble learning. Instead of using a particular combination of parameter values for anomaly detection, the proposed method generates the final result based on a set of results obtained using different parameter values. The experiments performed on datasets with known ground truth show that the proposed ensemble approach can outperform existing grammar-induction-based approaches with different criteria for selection of parameter values. We also show that the proposed approach, which has a linear time complexity with respect to the data size, can achieve performance similar to that of the state-of-the-art distance-based anomaly detection approach that has a quadratic time complexity. 

\begin{acks}
Sandia National Laboratories is a multimission laboratory managed and operated by National Technology and Engineering Solutions of Sandia, LLC, a wholly owned subsidiary of Honeywell International, Inc., for the U.S. Department of Energy’s National Nuclear Security Administration under contract DE-NA0003525.
\end{acks}

\bibliographystyle{ACM-Reference-Format}
\bibliography{sample-base}

%
% ...

\end{document}